\documentclass[preprint,12pt]{elsarticle}

\usepackage{lineno,hyperref}
\usepackage{bm}
\usepackage{amsfonts}
\usepackage{amssymb}
\usepackage{amsmath}
\usepackage{isomath}

\usepackage{adjustbox}

\usepackage{scalerel,stackengine}
\stackMath
\newcommand\reallywidehat[1]{%
\savestack{\tmpbox}{\stretchto{%
  \scaleto{%
    \scalerel*[\widthof{\ensuremath{#1}}]{\kern-.6pt\bigwedge\kern-.6pt}%
    {\rule[-\textheight/2]{1ex}{\textheight}}
  }{\textheight}%
}{0.5ex}}%
\stackon[1pt]{#1}{\tmpbox}%
}

\usepackage{stmaryrd} 
\usepackage[usenames]{color}
\usepackage{epsfig}
\usepackage{graphicx}
\usepackage{psfrag}

\graphicspath{{Figures/}}
\usepackage{float}
\floatstyle{boxed}
\newfloat{algorithm}{htb}{loa}
\floatname{algorithm}{Algorithm}
\usepackage{algorithm,algpseudocode}
\usepackage{caption}
\usepackage{subcaption}
\usepackage{color}
\usepackage{xcolor}
\usepackage{multirow}
\usepackage{natbib}
\modulolinenumbers[5]

\usepackage[left = 2cm, right=2.5cm, top=2cm, bottom =2cm]{geometry}

\begin{document}
\begin{frontmatter}

\title{ A finite operator learning technique for mapping the elastic properties of microstructures to their mechanical deformations }








\author{ Shahed Rezaei$^{1,*}$, Reza Najian Asl$^{2}$, Shirko Faroughi$^{3,*}$, Mahdi Asgharzadeh$^{3}$, Ali Harandi$^{4}$, \\Rasoul Najafi Koopas$^{5}$, Gottfried Laschet$^1$, Stefanie Reese$^4$, Markus Apel$^{1}$ }
\address{$^1$ACCESS e.V., Intzestr. 5, D-52072 Aachen, Germany}
\address{$2$ Chair of Structural Analysis, Technical University of Munich, Arcisstr. 21, 80333 München, Germany}
\address{$^3$Faculty of Mechanical Engineering, Urmia University of Technology, Urmia, Iran}
\address{$^4$Institute of Applied Mechanics, \\ RWTH Aachen University, Mies-van-der-Rohe-Str. 1, D-52074 Aachen, Germany}
\address{$^5$Institute of Solid Mechanics, Helmut-Schmidt University/University of the Federal Armed Forces, Holstenhofweg  85, Hamburg, 22043, Germany}
\address{$^*$ corresponding authors: s.rezaei@access-technology.de, sh.farughi@uut.ac.ir}

\begin{abstract}
\color{black}
To obtain fast solutions for governing physical equations in solid mechanics, 
we introduce a method that integrates the core ideas of the finite element method 
with physics-informed neural networks and concept of neural operators. 
This approach generalizes and enhances each method, 
learning the parametric solution for mechanical problems without relying on data 
from other resources (e.g. other numerical solvers).
We propose directly utilizing the available discretized weak form in finite element packages to construct the loss functions algebraically, thereby demonstrating the ability to find solutions even in the presence of sharp discontinuities. 
Our focus is on micromechanics as an example, where knowledge of deformation and stress fields for a given heterogeneous microstructure is crucial for further design applications. 
The primary parameter under investigation is the Young's modulus distribution 
within the heterogeneous solid system.
Our investigations reveal that physics-based training yields higher accuracy 
compared to purely data-driven approaches for unseen microstructures. 
Additionally, we offer two methods to directly improve the process of obtaining high-resolution solutions, avoiding the need to use basic interpolation techniques.
First is based on an autoencoder approach to enhance the efficiency for calculation on high resolution grid point. 
Next, Fourier-based parametrization is utilized to address complex 2D and 3D problems in micromechanics.
The latter idea aims to represent complex microstructures efficiently using Fourier coefficients.
Comparisons with other well-known operator learning algorithms, further emphasize the advantages of the newly proposed method.
\end{abstract} 

\end{frontmatter}
\color{black}

\newpage
\section{Introduction} 

By employing advanced simulation techniques, researchers can model and analyze the complex interplay of stresses, and deformations, providing thus a deeper comprehension of material properties. These simulations facilitate the exploration of diverse materials and contribute significantly to the design and optimization of materials with tailored mechanical characteristics, ultimately advancing innovation in numerous manufacturing industries. Examples of this include obtaining the mechanical behavior of the microstructure through numerical methods, which then, through homogenization, one obtains the material properties at the macro scale \cite{GEERS20102175}.

Notable examples of simulation techniques include finite difference, finite volume, and finite element methods (FEM) \cite{Faroughi2022, Liu2022}. Despite their predictive power, these methods have two major drawbacks. Firstly, obtaining the solution from them can easily become very time-consuming and they are not fast enough for many upcoming design applications. Secondly, as soon as any parameter changes (e.g., the morphology of the microstructure), one has to recompile the adapted model and redo the computation to obtain the new solution. In other words, the standard solvers are limited to one particular boundary value problem (BVP) and do not seem to be a sustainable and green choice, since they are designed to be used only once. 

Deep learning (DL) methods provide solutions to the stated problem by leveraging their robust interpolation capabilities and mapping disparate input and output spaces together. For an overview of the potential of deep learning methods in the field of computational material mechanics see \cite{Faroughi2022, Herrmann2024, Kim2024}.
The interpolation power of deep learning models is raised to a level that encourages researchers to train the neural network to learn the solution to a given boundary value problem in a parametric way. This idea is now pressed as operator learning, which involves the mapping of two infinite spaces or functions to each other. A major advantage is that once the training is done, the network evaluation for obtaining the solution is usually an extremely cheap process in terms of computational cost. Some well-established methods for operator learning include but are not limited to, DeepOnet \cite{Lu2021}, Fourier Neural Operator \cite{li2021fourier}, Graph Neural Operator \cite{li2020neural} and Laplace Neural Operator \cite{chen2023learning, cao2023lno}.

\subsection{Data-driven operator learning}
In this category, the data for the training is obtained from the available resources or by performing offline computations and/or experimental measurements for a set of parameters of interest. The idea can also be combined with different architectures of convolutional neural networks (CNN), recurrent neural networks, etc., depending on the application.

To further explore this topic, one can find comparative studies on available operator learning algorithms. \citet{LU2022114778} compared the relative performance and accuracy of two well-established operator learning methods (DeepONet and Fourier Neural Operator) for different benchmarks. \citet{RASHID2023105444} made a comparison between different neural operator architectures to predict strain in a data-efficient manner in 2D digital composites.

As with any other method, despite its benefits, there are potential drawbacks or new challenges that need to be investigated. Firstly, the cost of the training step and secondly their rather poor performance once we go beyond the range of training data and thirdly, be sure that any predictions of these algorithms are physically consistent or not. Therefore, researchers introduce physical constraints and expect to reduce the amount of data required to train the deep learning models. 

\subsection{Physics-driven operator learning}
To address the mentioned issues, another mainstream goes in the direction of integrating the model equations into the loss function of the neural network \cite{Lagaris1998, SIRIGNANO20181339, RAISSI2019}. Considering the concept of Physics-Informed Neural Networks (PINNs) as introduced by \citet{RAISSI2019}, the accuracy of predictions or network results can be significantly improved \cite{REZAEI2022PINN}. Additionally, in scenarios where the underlying physics of the problem is completely known and comprehensive, it becomes possible to train the neural network without any initial data. 
Among many contributions, for solid mechanics see \cite{HAGHIGHAT2021, REZAEI2022PINN, WU2023112521, ROY2023472}, and for constitutive material behavior see \cite{rezaei2023learning, HAGHIGHAT2023105828}. The PINN framework is also extended by using the energy form of the problem known as deep energy method (DEM) \cite{SAMANIEGO2020112790, NGUYENTHANH2020103874}, mixed formulations \cite{FUHG2022110839, REZAEI2022PINN}, domain decompositions \cite{JAGTAP2020113028, KHARAZMI2021113547}, and other techniques that enhance their predictability \cite{McClenny22, Faroughi2023, wang2023experts}. 
\textcolor{black}{In the above-mentioned methods, automatic differentiation is primarily utilized to construct the loss function based on available physical knowledge. The advantage is the so-called mesh-free approach, which can handle complex geometries. The drawback is that the process of automatic differentiation is costly compared to algebraic methods. Furthermore, one often has to adapt the location of collocation points to capture sharp transitions in the solution.
Finally, the two mentioned problems with the standard FEM method (i.e., computational cost and being limited to one BVP) still hold for new solvers such as PINN, DEM, and mixed-PINN. In fact, they still cannot compete with FEM for forward problems, although they have the potential to excel in inverse problems where identifying the underlying equations (e.g. parameter identification) is the primary concern.}
Therefore, an interesting trend and scientific question is how to combine ideas from operator learning and physics-informed neural networks.
\citet{li2023physicsinformed} combined training data and physical constraints in the context of Fourier neural operators to approximate the solution operator for some popular PDE families. \citet{Wang_Paris2021} proposed physics-informed DeepONets, which ensure the physical consistency of DeepONet models by leveraging automatic differentiation to impose physical laws via soft penalty constraints. The authors reported a significant improvement in the predictive accuracy of DeepONet models, and they also managed to reduce the need for large training datasets. 
See also \cite{ZHU201956, GAO2021110079, Liu2024} for utilizing CNN combined with physical constraints. In a study by \citet{ZHANG2021100220}, a physics-informed neural network tailored for analyzing digitalized materials was introduced, trained without labeled ground data using minimum energy criteria. For increased efficiency and potentially greater accuracy, \citet{kontolati2023learning} suggested mapping high-dimensional datasets to a low-dimensional latent space of salient features using suitable autoencoders and pre-trained decoders. \citet{ZHANG2023116214} presented encoder-decoder architectures to learn how to solve differential equations in weak form, capable of generalizing across domains, boundary conditions, and coefficients. 


While physics-informed operator learning shows promise, efficiently computing derivatives for input variables through automatic differentiation, especially for higher-order partial differential equations, remains a challenge. Furthermore, the presence of discontinuities in the solution makes the training of standard PINN models more challenging. A recent trend seeks to enhance efficiency by approximating derivatives using classical numerical methods, transforming the loss function from a differential equation to an algebraic one. However, this approach may introduce discretization errors, impacting the network's accuracy. \textcolor{black}{Note that the process of training the network using the backpropagation algorithm still relies on automatic differentiation.} Research conducted by \citet{REN2022114399}, \citet{ZHAO2023105516}, and \citet{ZHANG2023111919} investigated the potential of physics-driven convolutional neural networks utilizing finite difference and volume methods. 
\textcolor{black}{\citet{Fuhg2023} utilized CNN filters adjusted to resemble finite difference operators, performing minimization of energy functionals, which lowers the order of the differential operators compared to residual-based methods. \citet{He2023} employed a graph convolutional network in the deep energy method (DEM) model to handle irregular domains. Authors also showed that computation of the spatial gradient based on shape functions are more robust and delivers an accurate solution even at severe deformations.}


\color{black}
A summary of available solvers and surrogate models, along with their advantages and shortcomings is provided below to clarify our contributions. 
Standard solvers like finite element and finite difference methods, as well as approaches such as standard PINNs, deep energy methods, and mixed PINNs, are effective for solving specific PDEs in engineering applications. However, they are limited to single-use solutions for particular problems.

Neural operators aim to address this challenge by learning equations in a parametric way, often relying on heavy input data to effectively learn patterns. To circumvent the costly data generation step for such surrogate models, attention has turned to physics-informed neural operators to solve and provide solutions for a class of problems in a parametric manner. In this context, the term "solver" refers to algorithms capable of finding correct solutions without relying on data from external sources. However, this aspect, despite its importance, is still under heavy development, as training deep learning models efficiently without available data and making them flexible enough to handle sharp transitions in solution fields and complex geometries remains a challenge in many engineering applications.

To address such problems, we propose a new approach heavily motivated by reviewed works. Firstly, we suggest using the discretized formulation from FEM to construct the loss function. As a result, we can handle Neumann boundary conditions and complex geometries much more easier, while also eliminating the need for automatic differentiation in constructing the loss function, thereby increasing the efficiency of the training process.
\textcolor{black}{In this method, the physical degrees of freedom are listed as neurons in the output layer. After the training is done, the solution at any arbitrary point within the physical domain can be interpolated using the shape functions of the finite element method. In that sense, the method is similar to other neural operators such as DeepOnet, where infinite-dimensional mapping is achieved.}
Furthermore, we discuss the potential of Fourier-based deep learning models, allowing for the straightforward handling of much more complex topologies in 2D and 3D settings.

The proposed approach has the potential to be directly coupled with many available finite element-based solvers (or other solvers based on finite difference and spectral methods) in a way that it can directly use their residuals as the loss function, avoiding additional implementation efforts. We apply these techniques, named as finite operator learning, to investigate problems in micromechanics, particularly in the context of heterogeneous microstructures.
\color{black}

\begin{figure}[H] 
  \centering
  \includegraphics[width=0.99\linewidth]{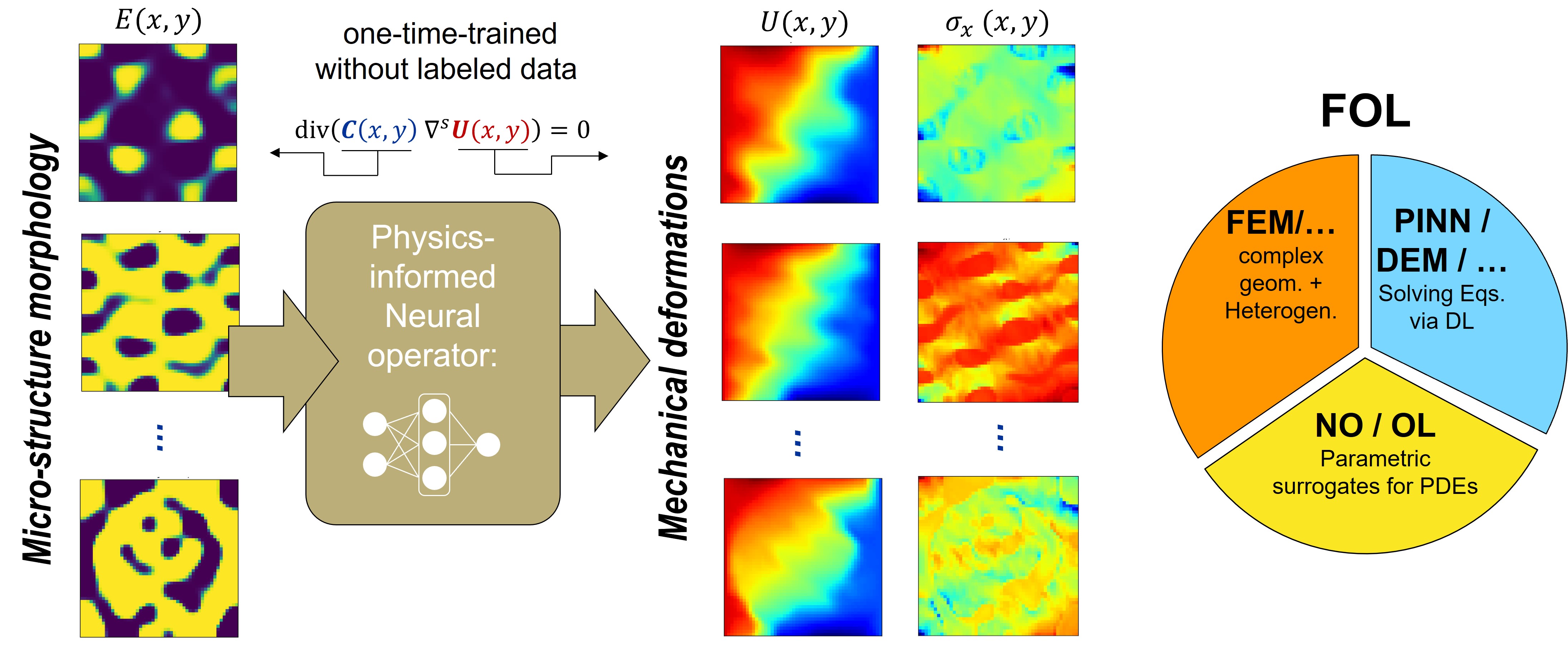}
  \caption{Left: By employing physics-informed operator learning, we aim to map the microstructure property to its mechanical deformation field under specified boundary conditions. \textcolor{black}{Right: the current method, coined as Finite Operator Learning (FOL), takes advantage of ideas from the Finite Element Method (FEM), Physics-Informed Neural Networks (PINN), deep energy method (DEM), and Neural Operators (NO). In particular cases, the FOL can represent each method individually.} }
  \label{fig:intro}
\end{figure}

Our approach maps parametric input field(s), namely elasticity parameters, to solution field(s), i.e., deformation components (see Fig.~\ref{fig:intro}). We aim to demonstrate that integrating ideas from the FEM to discretize the domain into finite pieces helps the network to find accurately and efficiently the proper mapping.

In Section 2, we review the problem formulation, which mainly deals with standard linear mechanics in materials having a heterogeneous microstructure. In Section 3, we focus on the application of deep learning to solve the discretized weak form in a parametric fashion, where the parameter under focus is the distribution of stiffness (i.e. Young modulus) within the microstructure. In Section 4, we present the results as well as additional insights on how to increase the performance and efficiency of the introduced method through the Fourier-based parameterization version. Finally, the paper concludes with discussions on potential future developments.

\color{black}
\color{black}

\section{Problem formulation}
For the mechanical problem in a 2D heterogeneous solid where the position of material points is denoted by $\bm{X}^T=[x,~ y]$. 
By denoting the displacement components by $U$ and $V$ in the $x$ and $y$ directions, respectively. 
Utilizing the standard finite element method, for each element the deformation field, stress tensor $\hat{\bm{\sigma}}$ as well as elasticity field $E, \nu$ are approximated as
\begin{align}
U &= \sum {N}_i u_i =\boldsymbol{N} \boldsymbol U_e, \\
V &= \sum {N}_i V_i =\boldsymbol{N} \boldsymbol V_e, \\
E &= \sum {N}_i E_i = \boldsymbol{N} \boldsymbol E_e, \\
\nu &= \sum {N}_i \nu_i = \boldsymbol{N} \boldsymbol \nu_e, \\
\hat{\bm{\sigma}} &= \bm{C}\boldsymbol{B} \bm{U}_e, 
\end{align}
Here, $\boldsymbol U^T_e=\left[
\begin{matrix}
u_1  &\cdots & u_4 \\
\end{matrix}
\right]$ and $\boldsymbol E^T_e=\left[
\begin{matrix}
E_1 &\cdots & E_4
\end{matrix}
\right]$ are the nodal values of the deformation field in the $x$ direction and elasticity of element $e$, respectively. Same holds for $\bm{V}_e$ and $\boldsymbol \nu_e$.  Moreover, matrices $\boldsymbol N$ and $\boldsymbol B$ store the corresponding to bi-linear shape functions and their spatial derivatives for a quadrilateral 2D element (see Appendix A for more details).
Following the standard procedure in the finite element method, one can write the so-called \textit{discretized} version of the weak form for one element as
\begin{align}
\label{eq:dis_residual}
\boldsymbol r_{e} = - \int_{\Omega_e} [\boldsymbol B]^T \boldsymbol C~[\boldsymbol B]\boldsymbol U_e ~dV + \int_{\Omega_t}[\boldsymbol N]^T \underbrace{\boldsymbol C~[\boldsymbol B \boldsymbol u_e]^T ~\boldsymbol n}_{\bar{\boldsymbol t}}~dS.
\end{align}
Here $\bar{\boldsymbol t}$ denotes the external traction on Neumann BCs and $\boldsymbol r_{e}$ represents the elemental residual vector which in this case is a $8 \times 1$ vector.

In the  current work, we are considering a combination of Neumann and Dirichlet boundary conditions, while periodic boundary conditions have not yet been addressed.

\section{A deep learning model to learn the static problem on an elastic heterogeneous domain}
\subsection{Neural network model}
This study relies solely on standard feed-forward neural networks where for each neural network we have the conventional architecture consisting of an input layer, possibly several hidden layers, and an output layer. All layers are interlinked, transmitting information to the subsequent layer. The computation of each component of the vector $\bm{z}^l$ is expressed as follows:\begin{equation}
\label{eq:NN_1}
    {z}^l_m = {a} (\sum_{n=1}^{N_l} w^l_{mn} {z}_n^{l-1} + b^l_{m} ),~~~l=1,\ldots,L. 
\end{equation}
The component $w_{mn}$ shows the weight between the $n$-th neuron of the layer $l-1$ and the $m$-th neuron of the layer $l$. Every neuron in the $l$-th hidden layer owns a bias variable $b_m^l$. The number $N_I$ corresponds to the number of neurons in the $l$-th hidden layer. The total number of hidden layers is $L$. The letter $a$ stands for the activation function.

\begin{figure}[H] 
  \centering
  \includegraphics[width=0.99\linewidth]{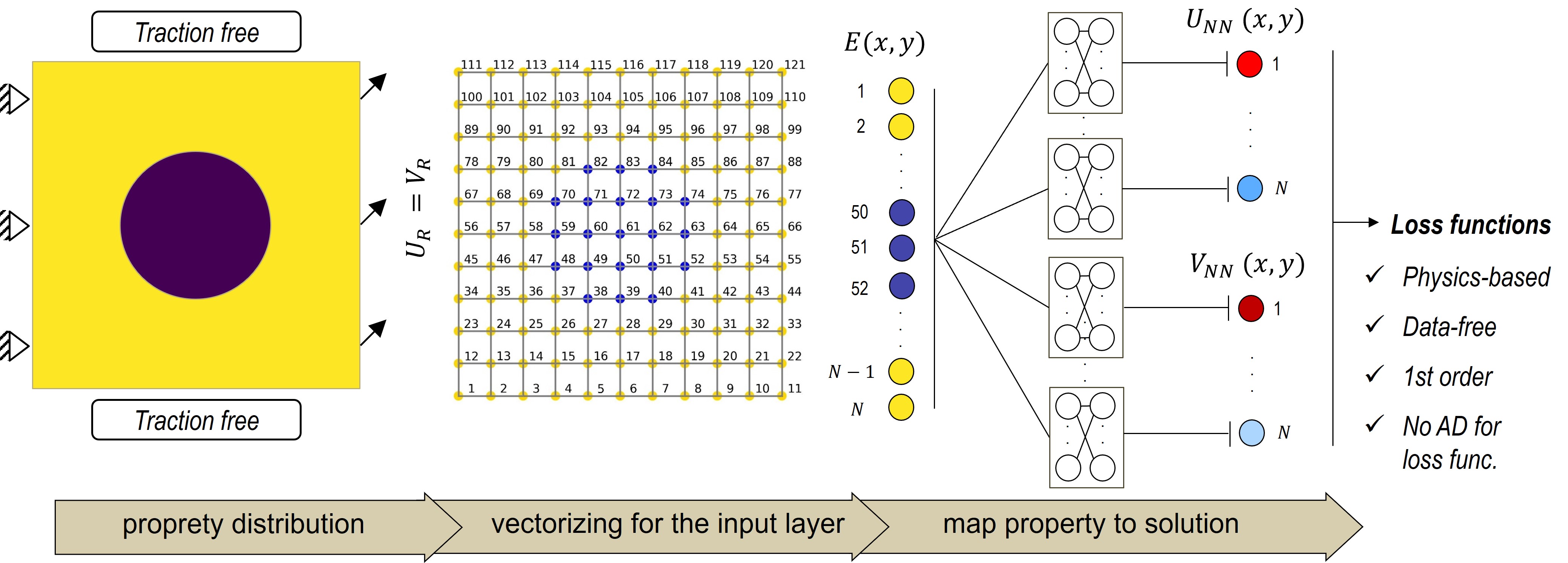}
  \caption{ Network architecture for finite operator learning (FOL), where information about the Young modulus distribution goes in and the unknown displacement components are evaluated utilizing a series of separate deep neural networks. For constructing the physical loss function, automatic differentiation (AD) is not required. }
  \label{fig:NN_idea}
\end{figure}

In a standard PINN framework \cite{RAISSI2019}, the network's input is the locations of the collocation points. In the current work of Finite Operator Learning (FOL) \cite{rezaei2024integration}, we utilize the so-called \textit{collocation fields}, which constitute randomly generated and admissible parametric spaces used to train the neural network. Here, collocation fields represent possible choices for the elastic properties, i.e. $E(x,y)$ and $\nu(x,y)$. \textcolor{black}{More details on how to generate samples are provided in section 3.3 and 4.5.}
Therefore, the input layer consists of information on Young modulus $\{E_j\}=\{E_1,\cdots, E_{N}\}$ and Poisson coefficient values $\{\nu_j\}=\{\nu_1,\cdots, \nu_{N}\}$ at all the nodes, and the output layer consists of the components of the mechanical deformation at each discretization node $i$. i.e. $\{U_i\}=\{U_1,\cdots, U_{N}\}$ and $\{V_i\}=\{V_1,\cdots, V_{N}\}$. 
The neural network model is therefore summarized in the following steps which are also depicted on the right-hand side of Fig.~\ref{fig:NN_idea})
\begin{align}
\label{eq:in_out}
    \bm{X} &= [\{E_j\},~\{\nu_j\}],~~~\bm{Y} = [\{U_i\},~\{V_i\}],~~~i,j = 1 \cdots N, \\
    U_{i} &= \mathcal{N}_{U_i} (\bm{X}; \bm{\theta}_{U_i}),~~~V_{i} = \mathcal{N}_{V_i} (\bm{X}; \bm{\theta}_{V_i}),~~~\bm{\theta}_i = \{\bm{W}_i,\bm{b}_i\}.
\end{align}
Here, the trainable parameters of the $i$-th network (i.e. $\bm{W}_i$ and $\bm{b}_i$) are denoted by $\bm{\theta}_i$. 
Note that we propose employing separate fully connected feed-forward neural networks for each output variable. The number of nodes is denoted by $N$, set to $121$ in this paper.

\textcolor{black}{The domain remains square-shaped. However, the formulation is very general and can be easily adapted for irregular geometries and unstructured meshes, thanks to the power of finite element techniques \cite{yamazaki2024finite}.}
Note that in Fig.~\ref{fig:NN_idea} and throughout the rest of the work, we assume that the Poisson ratio remains constant in the RVE, and our focus lies solely on the variation of Young's modulus. The latter assumption helps to simplify the input parameter space but does not constrain the methodology. 
Next, we introduce the loss function. We assume traction-free boundary conditions on the top and bottom edges, resulting in vanishing traction at the Neumann boundaries. The displacement components at the left edge are fixed (i.e., $U_L = V_L = 0$). On the right edge, we apply tension and shear simultaneously (i.e., $U_R = V_R = 0.05~$mm on Fig.~\ref{fig:NN_idea}.) leading to a rather general Dirichlet boundary condition.


The total loss term $\mathcal{L}_{\text{tot}}$ combines both the elemental energy form of the governing equation $\mathcal{L}_{\text{en,e}}$ and the Dirichlet boundary terms $\mathcal{L}_{\text{db,i}}$. It is important to note that, thanks to the weak formulation, the Neumann boundary conditions are automatically included in $\mathcal{L}_{\text{en,e}}$. Following the approach outlined in \cite{rezaei2024integration} for thermal problems, the total loss function involves the integration of element residual vectors (i.e., Eq.~\ref{eq:dis_residual}) using Gaussian integration, resulting in
\begin{align}
\label{eq:loss_tot}
\mathcal{L}_{tot} &= \sum_{e=1}^{n_{el}}\mathcal{L}_{en,e}(\boldsymbol \theta) + \sum_{i=1}^{n_{db}} \mathcal{L}_{db,i}(\boldsymbol \theta), \\
\label{eq:loss_ene}
\mathcal{L}_{en,e} &= \boldsymbol U^T_e(\boldsymbol \theta) \left[ \boldsymbol K_e~\boldsymbol U_e(\boldsymbol \theta) \right], \\
\label{eq:loss_db}
\mathcal{L}_{db,i} &= \dfrac{1}{n_{db}} |U_i(\boldsymbol \theta) - U_{i,db}| +
\dfrac{1}{n_{db}} |V_i(\boldsymbol \theta) - V_{i,db}|,\\
\boldsymbol K_{e} &= \sum_{n=1}^{n_{int}} \dfrac{w_n}{2}~\text{det}(\boldsymbol J)~[\boldsymbol B(\boldsymbol \xi_n)]^T \boldsymbol C_e(\boldsymbol \xi_n) \boldsymbol B(\boldsymbol \xi_n), \\
\boldsymbol{C}_e(\boldsymbol \xi_n) &=  \dfrac{\boldsymbol{N}(\boldsymbol \xi_n) \boldsymbol E_e}{1-\nu^2}\,\begin{bmatrix} 1 & \nu & 0 \\
\nu & 1 & 0 \\
0 & 0 & \dfrac{1-\nu}{2}
\end{bmatrix}.
\end{align}
Here, $n_{\text{int}}=4$ represents the number of Gaussian integration points. Additionally, $\boldsymbol \xi_n$ and $w_n$ denote the coordinates and weighting of the $n$-th integration point and the determinant of the Jacobian matrix is denoted by $\text{det}(\boldsymbol J)$. See also Appendix A.
In Eq.~\ref{eq:loss_db}, $n_{\text{db}}=22$ denotes the number of nodes along the Dirichlet boundaries (i.e., the left and right edges). Additionally, $U_{i,\text{db}}$ and $V_{i,\text{db}}$ represent the given displacement at node $i$. The final loss function becomes
\begin{align}
\label{eq:loss_sum}
\boxed{
\mathcal{L}_{tot} =  
a_e
\sum_{e=1}^{n_{el}}\boldsymbol U^T_e \left[ \boldsymbol \sum_{n=1}^{n_{int}}~\boldsymbol B^T \boldsymbol C_e \boldsymbol B \right] \boldsymbol U_e  
+ 
a_b\sum_{i=1}^{n_{db}} |U_i - U_{i,db}| + |V_i - V_{i,db}|}.
\end{align}
In the above relation, the determinant of the Jacobian matrix is constant since we use quadrilateral elements. \textcolor{black}{Additionally, for the four integration points, we have $w_n=1$. Based on the energy formulation for the first term in Eq.~\ref{eq:loss_sum}, one can write $a_e=\dfrac{w_n}{2}\text{det}(\boldsymbol J)$. For the second term, we suggest $a_b=\dfrac{A}{n_{\text{db}}}$, which needs to be adapted based on the problem via a parameter study or novel adaptive weighting techniques. Note that this term can be limited by applying a hard boundary constraint.}
In Eq.\,\ref{eq:loss_sum}, $a_b$ represents the weighting of the boundary terms which guarantees the enforcement of the boundary conditions. Note that in Eq.\,\ref{eq:loss_sum}, we utilized the discretized energy form, but one can also use the norm of the total residual vector.
The total loss term $\mathcal{L}_{tot}$ is minimized with respect to the network parameters $\bm{\theta}$ concerning a batch of collocation fields.

\subsection{Short review on the DeepONet model}
\label{DeepONet}
\begin{figure}[H] 
  \centering
  \includegraphics[width=0.99\linewidth]{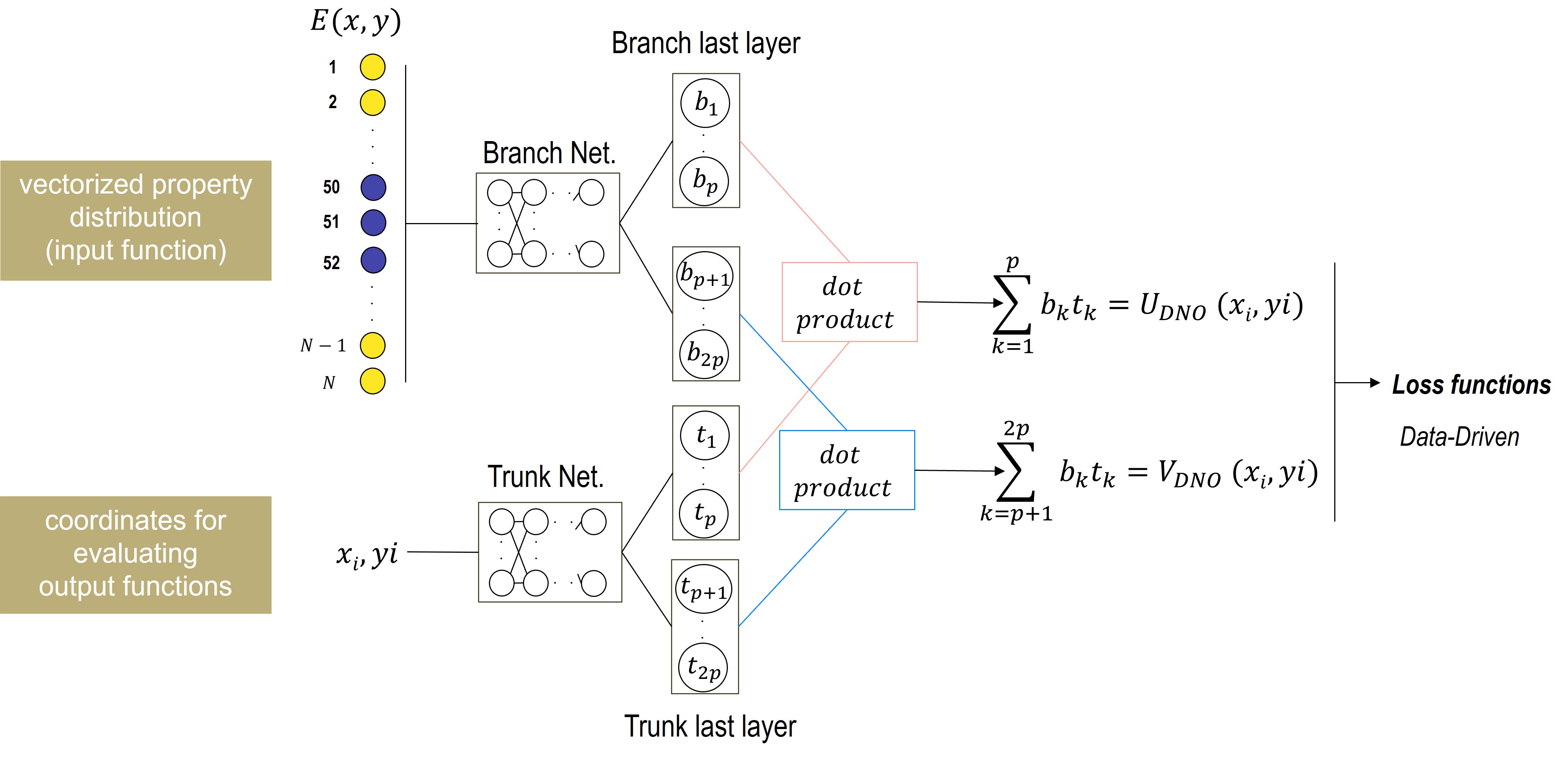}
  \caption{DeepONet's architecture for translating property maps into mechanical deformations. The property map is vectorized and sampled at nodes (acting as sensors) to serve as input to the branch network, while the trunk network receives coordinates for output evaluation. Both networks may have different hidden layers, but end up with the same number of neurons in their final layers, which are divided into two equal-length vectors. The dot product of these vectors produces the desired outputs ($U_i$ and $V_i$).}
  \label{fig:DNOarch} 
\end{figure}

Building on the foundational principles of the universal approximation theorem \cite{Chen1995UniversalAT}, the so-called vanilla DeepONet architecture \cite{li2020neural} with two outputs is used here. The network architecture consists of a single branch and single trunk network as depicted in Fig.\,\ref{fig:DNOarch}.

To obtain multiple outputs from a single DeepONet, we divide the final layers of the branch and trunk networks into two parts of equal length (see also \cite{wang2022improved}). Two operators that map the property distribution to the mechanical deformations are written as
\begin{equation}
\begin{aligned}
\label{eq:deeponetoperators}
\mathbf{G}_{\bm{\theta}}^{(U)}(\bm{E})(\bm{X}) &= \sum_{k=1}^{p} b_k\left(E(x_1, y_1), E(x_2, y_2), \ldots, E(x_{N}, y_{N})\right) \cdot t_k(\bm{X}), \\
\mathbf{G}_{\bm{\theta}}^{(V)}(\bm{E})(\bm{X}) &= \sum_{k=p+1}^{2p} b_k\left(E(x_1, y_1), E(x_2, y_2), \ldots, E(x_{N}, y_{N})\right) \cdot t_k(\bm{X}).
\end{aligned}
\end{equation}

In Eq.,\ref{eq:deeponetoperators}, $\mathbf{G}^{(U)}$ and $\mathbf{G}^{(V)}$ denote the solution operators for displacements in the $x$ and $y$ directions, respectively. 
Finally, for the pure data-driven DeepONet, the total loss function is written as
\begin{align}
\label{eq:deeponetloss}
\mathcal{L}_{DeepONet}\,&=\dfrac{1}{I_{train}\,N}\,\sum_{i=1}^{I_{train}}\,\sum_{j=1}^{N}|\mathbf{G}_{\bm{\theta}}^{(U)}(\bm{E}^{(i)})(\bm{X}_j^{(i)})-U^{(i)}_j|^2\,+\,|\mathbf{G}_{\bm{\theta}}^{(V)}(\bm{E}^{(i)})(\bm{X}_j^{(i)})-V^{(i)}_j|^2.
\end{align}
For more details about the DeepONet training see Appendix B. It's worth noting that several enhancements have been introduced to improve the performance of the original DeepONet architecture \cite{wang2022improved, haghighat2024deeponet}. 

\subsection{Preparation of collocation fields}
\label{preparationofcollocationfields}
The neural network mentioned above should be trained using initial input samples representing the Young modulus distribution. It is important to note that the training process is entirely unsupervised as no FEM solution is required and random fields are sufficient to initiate the training by having the loss function based on discretized FEM residual vector.

In the current work, we consider two-phase heterogeneous materials, which is a common example in many alloys and composite materials \cite{LASCHET2022143125, VOGIATZIEF2022165658}. The goal is to modify the microstructure's morphology and properties, allowing for arbitrary shapes and volume fraction ratios between the two phases. However, the elastic properties (i.e., Young's modulus and Poisson ratio) of each phase remain fixed. Additionally, to reduce complexity, we shall investigate at first only one ratio between the elastic properties of each phase. 

So far, we have the following model simplifications: keeping the Poisson ratio and phase contrast ratio constant, and using a grid of 11 by 11. Even with the previously mentioned simplifications, there can be approximately $2^{121} \approx 2.6 \times 10^{36}$ independent samples. Creating all these samples is unfeasible and may be unnecessary depending on considered application. Hence, the user should determine the relevant parameter space, specific to each problem, thereby preventing unnecessary complications. The latter point can be a challenging aspect in designing a very general deep learning model that works for all cases.

\begin{figure}[H] 
  \centering
  \includegraphics[width=0.95\linewidth]{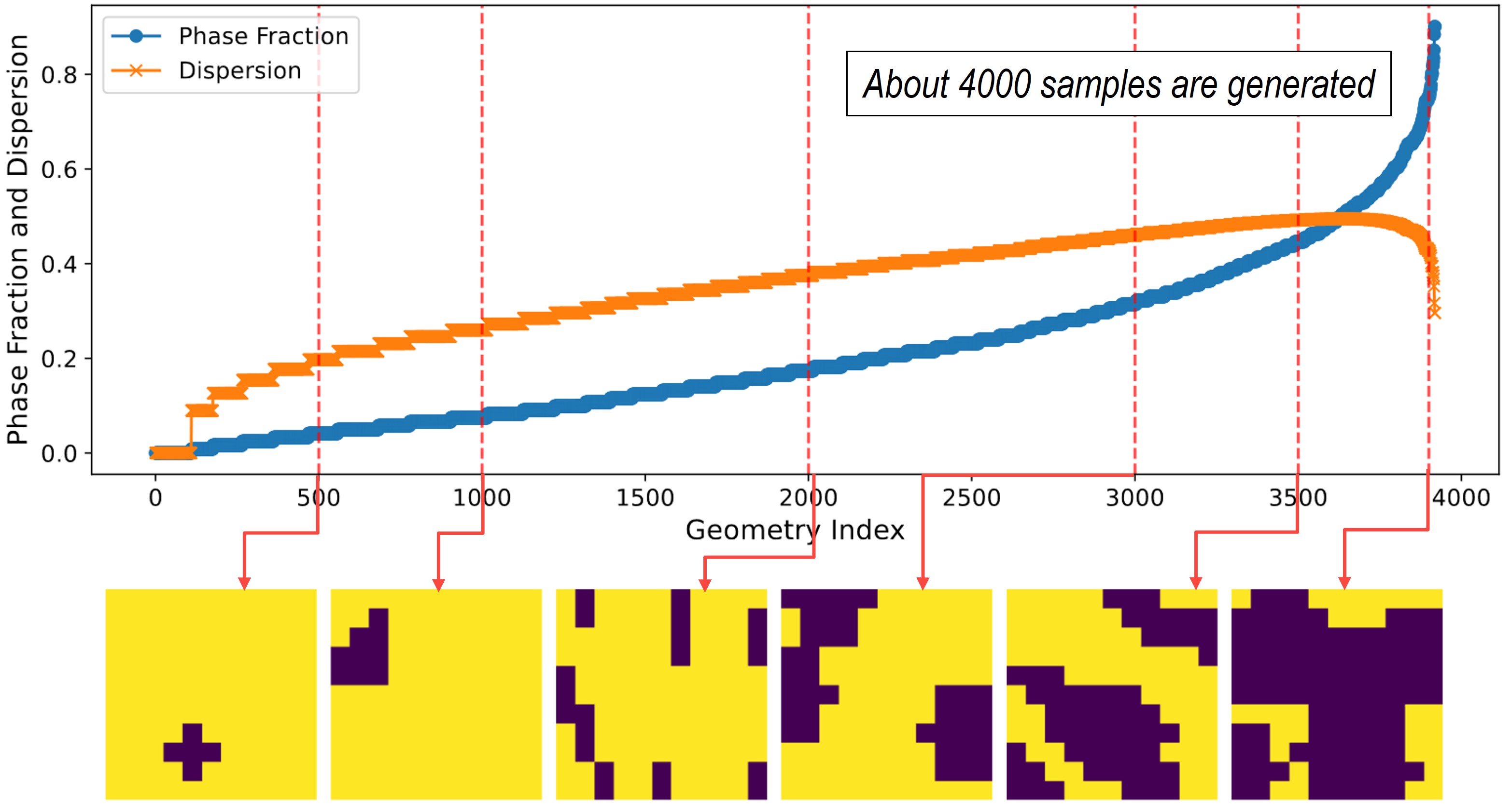}
  \caption{Examples showcasing created morphologies for a two-phase microstructure.}
  \label{fig:samples}
\end{figure}

As mentioned, our preliminary motivation relies on two-phase metallic materials, and after some investigations, we have created simply a rather meaningful set of samples for training \cite{rezaei2024integration}. We generated $4000$ samples for a grid of size $\sqrt{N}$ by $\sqrt{N}$, with $\sqrt{N}=11$. On Fig.~\ref{fig:samples}, where the generated microstructures are sorted based on their phase fraction value, six samples with low and high phase fraction values are shown. 

\begin{figure}[H] 
  \centering
  \includegraphics[width=0.9\linewidth]{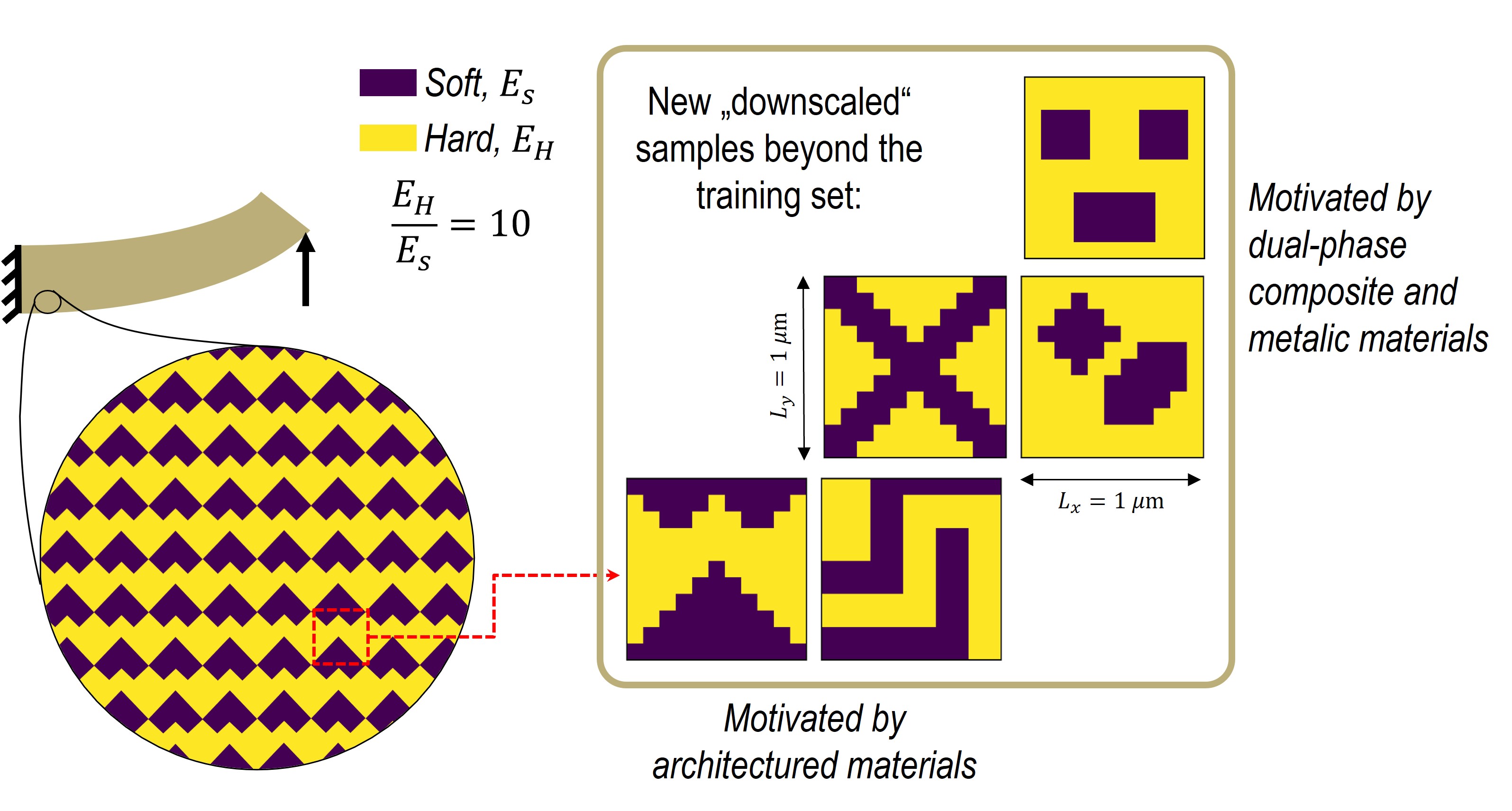}
  \caption{Unseen samples used for testing the neural network after training.  }
  \label{fig:test}
\end{figure}


To assess the NN's performance, additional samples are chosen for testing, five of them are shown in Fig.~\ref{fig:test}. These samples are entirely novel and deliberately selected for their symmetrical aspects, ensuring they are beyond the scope of the training dataset \textcolor{black}{(see also Fig.~\ref{fig:sample_1} Appendix C for an overview on the topology of the samples)}. These samples are motivated by dual-phase architectures in composite materials, where a soft and hard phase are combined. Note that all the samples are downsized to be represented by an 11 by 11 grid. The downsampling strategy is also described in \cite{rezaei2024integration}, where a simple CNN algorithm combined with a max-pooling is used \cite{mianroodi2022lossless}.


\section{Results}
The algorithms developed in this study are implemented \textcolor{black}{using JAX \cite{jax2018github} software. The codes are also implemented in SciANN package \cite{SciANN}, and the methodology can be adapted to other programming platforms as well.} A summary of the network's (hyper)parameters is presented in Table~\ref{tab:NN_para}. Note that whenever multiple options for a parameter are introduced, we examine the influence of each parameter on the results.
\textcolor{black}{Training can also be done on personal laptops without relying on GPU acceleration, thanks to the efficiency of the proposed framework. Nevertheless, we primarily used a Quadro RTX 6000 GPU unless otherwise noted.} 
\begin{table}[H]
\centering
\caption{Summary of the network parameters.}  
\label{tab:NN_para}
\begin{footnotesize}
\begin{tabular}{ l l }
\hline
Parameter                          &  Value    \\
\hline
Inputs, Outputs                  &  $\{E_j\}=\{E_1,\cdots, E_{N}\},~[\{U_i\}=\{U_1,\cdots, U_{N}\},~\{V_i\}=\{V_1,\cdots, V_{N}\}]$ \\ 
Activation function                &  Swish \cite{ramachandran2017searching} \\ 
Layers and neurons for each sub-network   &  $(L, N_l) = (2, 10)$ \\
Optimizer                          &  Adam \cite{kingma2017adam} \\ 
(Number of samples, Batch size)   &  $(100,2)$, $(1000,25)$, $(2000,50)$, $(4000,100)$ \\
(Learning rate $\alpha$, Number of epochs)  &  $(5 \times 10^{-4},4000)$ \\ 
\hline
\end{tabular}
\end{footnotesize}
\end{table} 


The material parameters listed in Table \ref{tab:par} do not necessarily represent a specific material at this point. For the chosen Young modulus values and boundary conditions we did not see the necessity to perform any additional normalization before the training. 
\begin{table}[H]
\centering
\caption{Model input parameters for the mechanical problem. }  
\label{tab:par}
\begin{footnotesize}
\begin{tabular}{ l l }
\hline
      &  Young modulus value/unit \\
\hline
Phase 1 ($E_{H}$)  & $1.0$~MPa  \\
Phase 2 ($E_{S}$)  & $0.1$~MPa  \\
Applied displacement, small defo. model ($U_R$,$V_R$)  & ($0.05$~mm, $0.05$~mm)\\
\hline
\end{tabular}
\end{footnotesize}
\end{table}

\subsection{Evolution of loss function and network prediction} 

\begin{figure}[H] 
  \centering
  \includegraphics[width=0.9\linewidth]{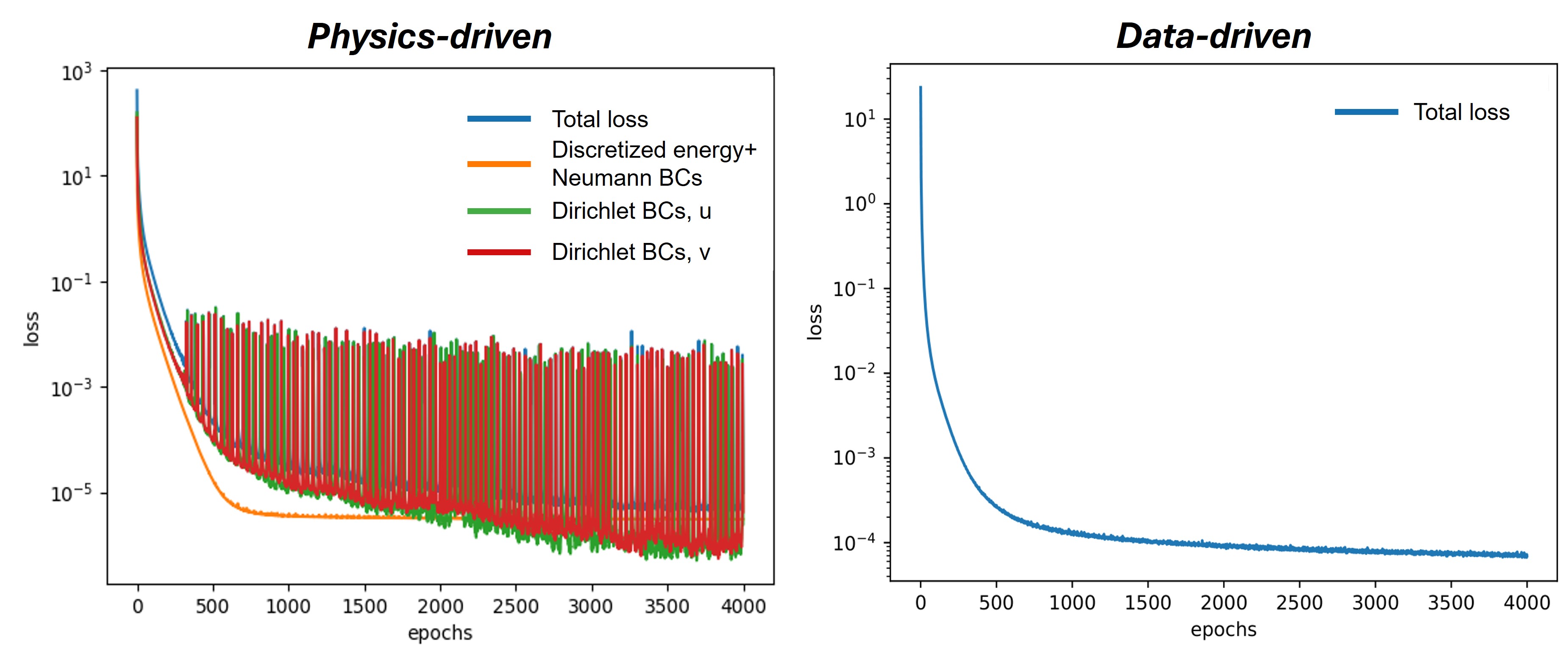}
  \caption{Evolution of loss terms for parametric learning of the quasi-static mechanical equilibrium problem.  }
  \label{fig:loss}
\end{figure}

The evolution of each loss term is illustrated in Fig.\,\ref{fig:loss} throughout epochs. On the left-hand side, the loss terms associated with the physics-driven model are presented, while the right-hand side displays the loss evolution for the supervised data-driven model. It is noteworthy that satisfactory results are achieved only after $4000$ epochs in both cases. It is important to mention that, for the data-driven model, offline finite element calculations are performed on the same set of 4000 samples, and the identical network architecture is employed as in the physics-driven model.

In Fig.\,\ref{fig:loss}, we employ the "swish" activation function, and 4000 samples with a batch size of 100 are used for training while the remaining parameters align with those in Table~\ref{tab:NN_para}. All the loss functions decay simultaneously, and we did not observe any significant improvement after $4000$ epochs.

For the loss term related to the Dirichlet boundary conditions, we assigned higher weightings to satisfy the boundary terms. Following a systematic study that began with equal weightings for both loss terms, we selected $A=10~n_{nb}$, indicating $a_b=10$. Choosing equal or lower weightings resulted in less accurate predictions within the same number of epochs as the boundary terms were not fully satisfied. Opting for higher values did not significantly improve the results. It is worth mentioning that with the current approach, one can also easily apply hard boundary constraints by omitting nodes (i.e., degrees of freedom) with Dirichlet boundary conditions from the output layers. We intentionally did not perform this for the current study to demonstrate that even with simple soft constraints, one can train the neural network properly. This aspect is crucial for extending the neural network to learn different boundary conditions in a parametric way for future work.

\textcolor{black}{In Figure \ref{fig:test1} and similar ones in Appendix C (Figs.~\ref{fig:test0} and \ref{fig:test2}), the predictions of the same neural network for various microstructures are depicted.} In the middle part of the figures, we display the reference solutions obtained by the FEM. All the results are assessed on finer meshes ($100 \times 100$) using bi-linear shape function interpolations. On the right-hand side, the absolute value of the point-wise difference between FOL and FEM is shown. All displacement values are denoted in micrometer $[\mu$m], stress components are represented in [GPa], and Young modulus is expressed in [GPa] too.
\begin{figure}[H] 
  \centering
  \includegraphics[width=1.0\linewidth]{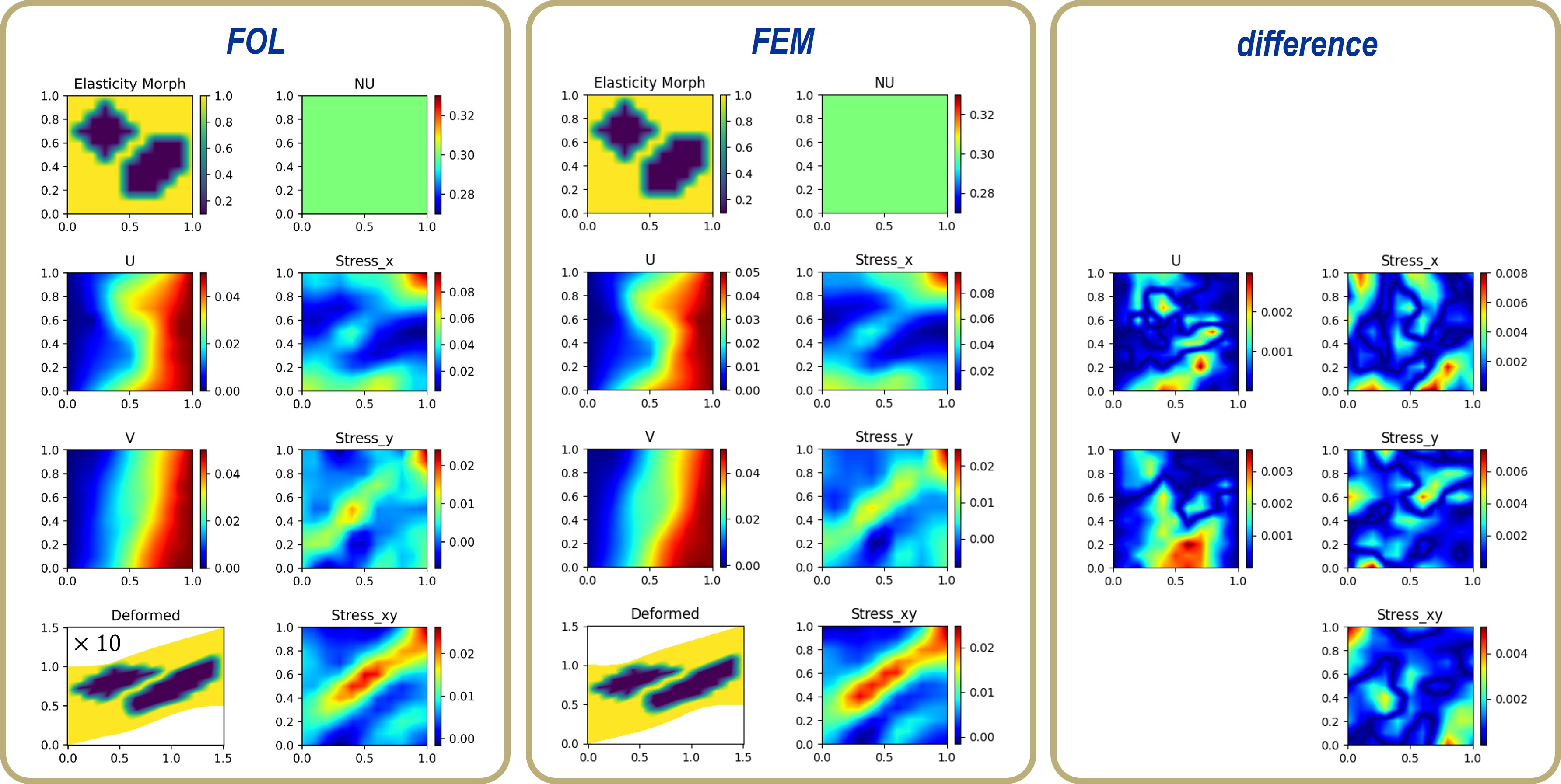}
  \caption{Predictions of the trained FOL for approximately circular and ellipsoidal inclusions.}
  \label{fig:test1}
\end{figure}

The predictions of the neural network for unseen microstructures are in very good agreement with those from the FEM. The maximum pointwise error for displacement components remains below $10\%$, localized only in some points of the entire domain. For the stress components, which are obtained through post-processing of deformations values, the errors are higher, as expected, since the neural network predictions are solely based on deformations, and any higher-order derivatives are more sensitive to slight deviations in the original solution. On the other hand, for many applications in multiscale material modeling, homogenized values of stress and strain components are needed, and these are well-predicted as outlined in Section 4.2. 

\textcolor{black}{Based on the reported results, we observe a tendency in the error distribution depending on the material topology. The main critical regions with higher errors are localized close to the phase boundaries where we expect a jump in the solution. A more quantitative report on the errors and possible remedies to decrease them is discussed in the following sections. In particular, we will discuss the influence of initial sample fields, using a pre-trained autoencoder, as well as a Fourier-based parameterization to reduce these errors.}

\subsection{The effect of the number of initial samples}
\textcolor{black}{In Fig.~\ref{fig:FOL_conv_DD_1} (and Fig.~\ref{fig:FOL_conv_DD_2} in Appendix C), we not only compare the performance of the data-driven and physics-driven FOL models but also study the influence of the initial training samples.} 

Note that other (hyper)parameters of the NNs are kept the same for this study (see also Tab.\,\ref{tab:NN_para}). As expected, reducing the number of samples from the initial dataset leads to increased errors. 

What is noteworthy from such observations is how accurately the physics-driven model can predict the solution even when trained based on $2000$ initial random samples. This confirms that providing equations for training has a more significant impact on the network performance. To confirm this observation, we tested several cases with different complexities in the next section.

\begin{figure}[H] 
  \centering
  \includegraphics[width=0.99\linewidth]{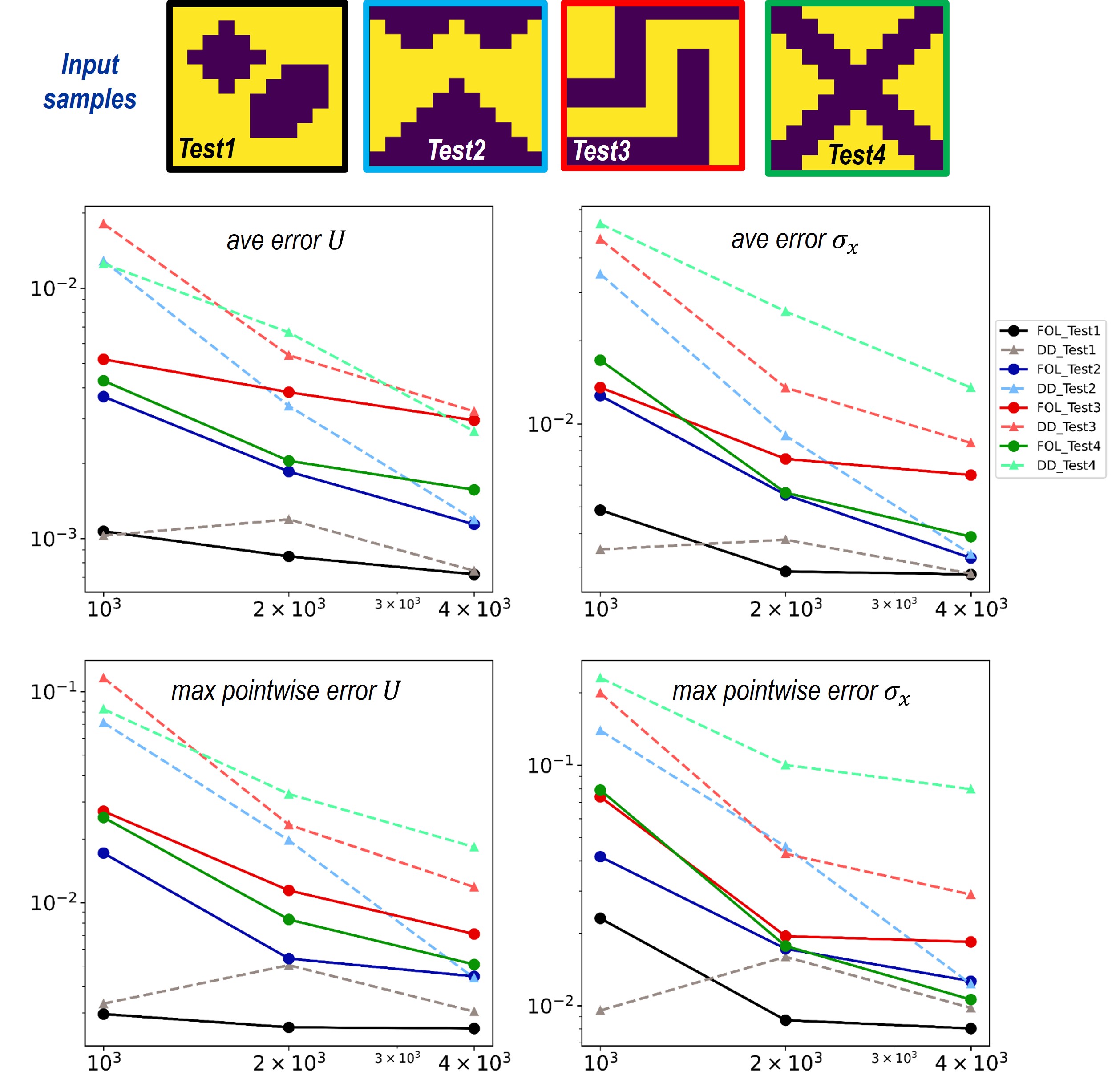}
  \caption{ Improving the performance of the deep learning models by increasing the number of randomly generated samples for training. Different error evaluations are utilized based on different applications, where the physics-driven model always behaves better than the data-driven one for the same unseen test cases.
  }
  \label{fig:fin_err}
\end{figure}

In Fig.~\ref{fig:fin_err}, we present quantitative error measurements by using different error metrics for four distinct unseen microstructure. Error metrics are defined as: $Err_{MSE} = \sqrt{\frac{1}{N}\sum_i (u_{NN}(i)-u_{FE}(i))^2}$ or as $Err_{MAX} = \text{Max}(|u_{NN}(i)-u_{FE}(i)|)$. The former indicates the average pointwise error, while the latter focuses on the maximum local difference. Training the model with more samples enhances its performance, and one can observe a convergence behavior similar to that of other numerical methods, despite are based on a different concept here.

From the comparison of error plots, two interesting points emerge. Firstly, the errors for the physics-driven model are generally lower than those for the data-driven model, regardless of the number of sample fields used for training the model. Secondly, for test case 1, which is more similar to the samples used for training, the errors of the data-driven model are very acceptable. This indicates that, apart from the quantity of the samples, the so-called quality of them plays a role. In other words, even if it sounds rather unfeasible to train the model for any possible distribution of Young modulus, one can always select a meaningful set of samples for each specific application to train the model properly. 
For test cases 3 and 4, we observe higher errors with both error metrics, as they are significantly distant from the sample shapes considered during training. Therefore, we expect that by further enriching the initial sample fields, the errors for these two cases will also decrease. 

\begin{figure}[H] 
  \centering
  \includegraphics[width=0.99\linewidth]{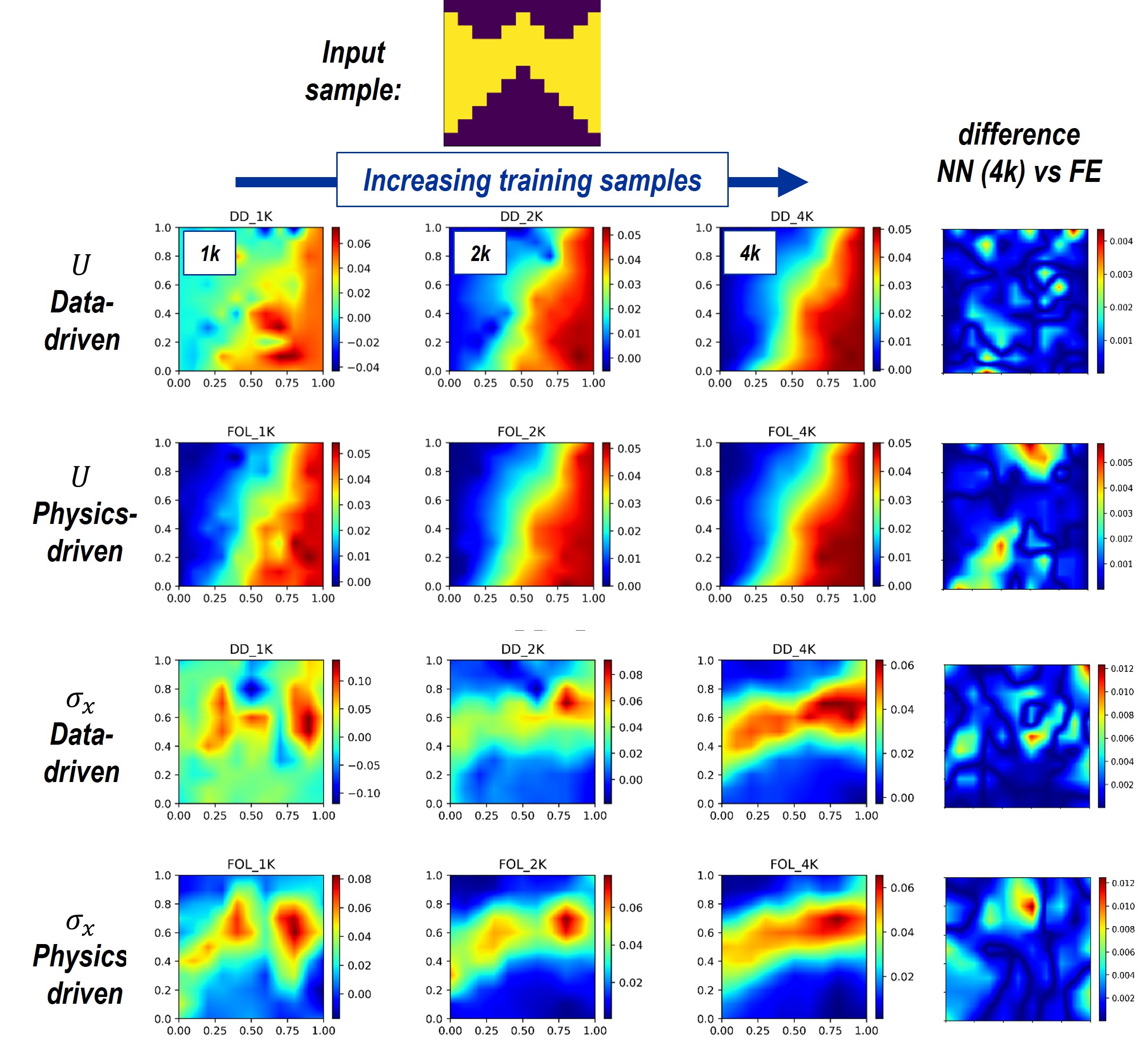}
  \caption{Performance of data-driven and physics-driven approaches for the different number of training samples.  }
  \label{fig:FOL_conv_DD_1}
\end{figure}

\subsection{Comparison with other neural operator techniques}

\begin{figure}[H] 
  \centering
  \includegraphics[width=0.99\linewidth]{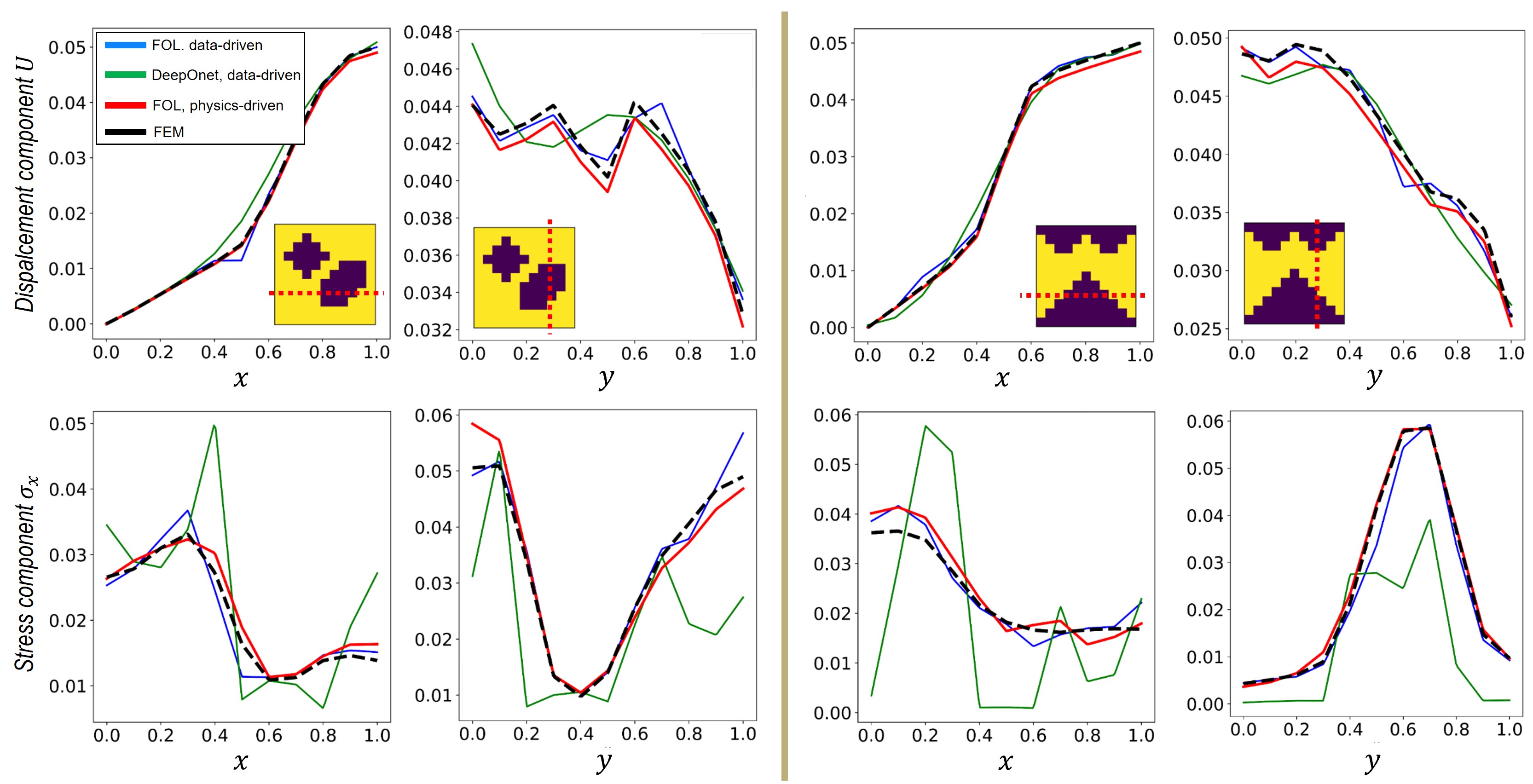}
  \caption{Comparison of data-driven FOL, data-driven DeepOnet, and physics-driven FOL approaches regarding the displacement component in x direction $U$ as well as stress component $\sigma_x$ along $x$ and $y$ directions, respectively.  }
  \label{fig:cross_plot}
\end{figure}

In this section, we compare the results of the introduced FOL method against one particular data-driven operator learning method, DeepONet. For a fair comparison, the DeepOnet model explained in Sec.\,\ref{DeepONet} is trained by using 4000 samples, the same data set discussed in the last section.
The comparison results, illustrated in Fig.\,\ref{fig:cross_plot} for several unseen test cases, reveal that the physics-driven method provides more accurate predictions for mechanical deformation as well as stress distribution.
\textcolor{black}{Note that the DeepOnet performance on predicting the displacement values is still acceptable to some extent, as DeepOnet is designed to achieve this goal. The stress values are obtained by taking the derivatives of the displacement field after training and there are no additional loss terms for them. This is one of the main reasons behind the larger deviations in predicting the stress values, as any fluctuations in the displacement predictions heavily influence the stress values.}

This highlights the attractiveness of the physics-driven FOL for two main reasons: Firstly, it operates without the need for labeled data during training, and secondly, it provides more accurate predictions for unseen cases for both displacement components (as primary variables) and their spatial derivatives (i.e., stress components). Nevertheless, it is essential to emphasize that supervised training incurs a lower training cost, averaging approximately $7$ to $10$ times less compared to the physics-based approach.

\textcolor{black}{The observations discussed above are not limited to a specific test case; they consistently appear across other unseen cases and regions as well.
It is noteworthy that in future works, other operator learning approaches should be properly compared under the same data/sample set to enable a fair comparison. Moreover, this reporting is done solely in  the case of linear elastic behavior of heteregoneous microstructures, and perhaps in presence of some nonlinearities, these conclusions may change. Therefore, more studies are required for a fair comparison between different neural operators. }

\subsection{Influence of phase contrast on the results}

\begin{figure}[H] 
  \centering
  \includegraphics[width=0.8\linewidth]{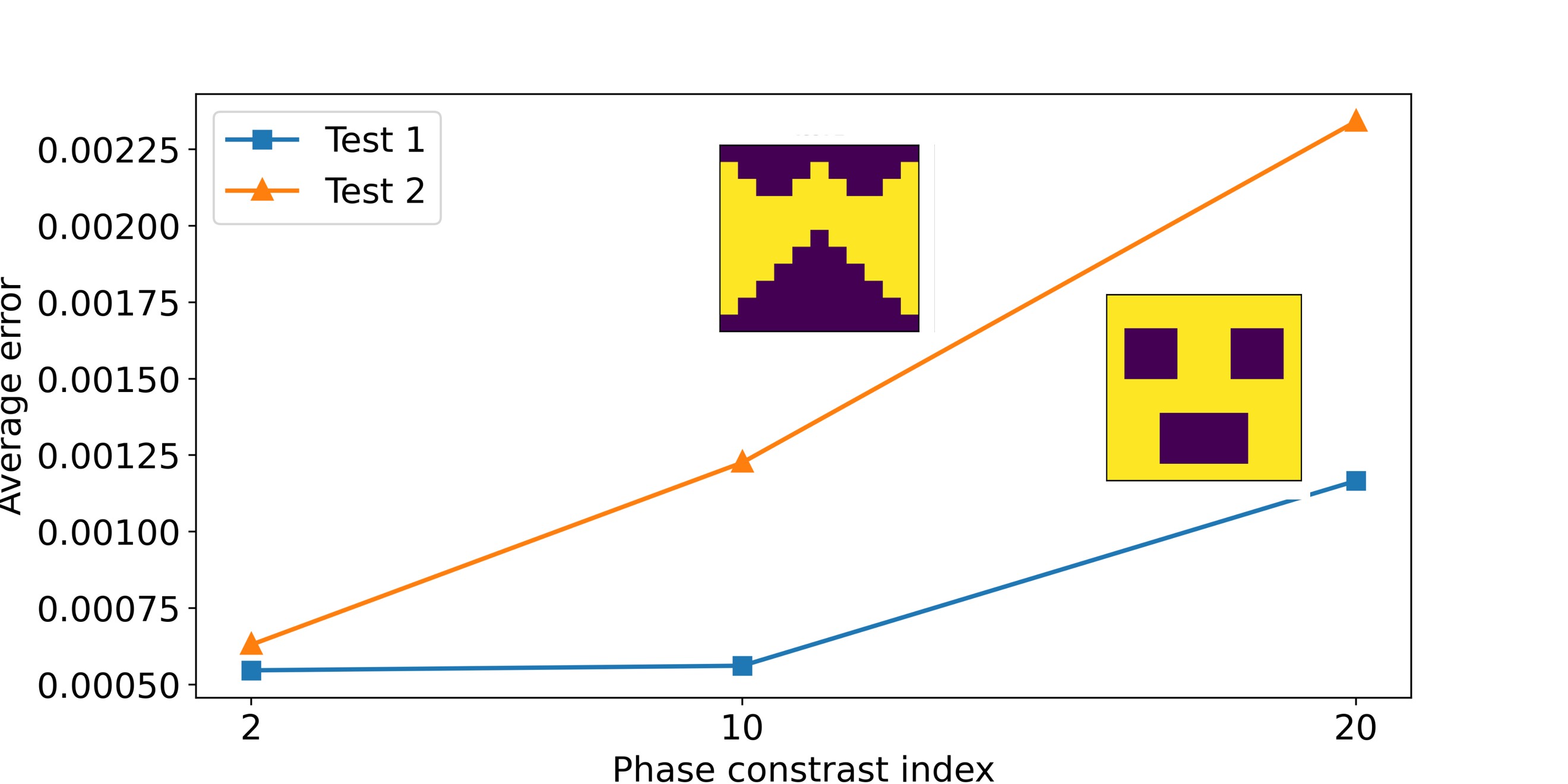}
  \caption{ Influence of ratio of phase contrast on the obtained results.
  }
  \label{fig:ratio}
\end{figure}

In this section, we briefly discuss the changes in the phase contrast (ratio of the Young modulus) based on the obtained results from the FOL. So far, we mainly focused on the ratio of $E_{H} / E_{S} = 10$. In Fig.~\ref{fig:ratio}, we also showed the averaged error for the displacement component $U$ in the $ x $ direction as we increase the stiffness ratio to $ E_{H} / E_{S} = 20 $  or decrease to $E_H/E_S = 2$. These latter values are roughly representative of some engineering applications in composite and dual-phase metallic materials.

We can conclude that by increasing the ratio, the errors generally tend to slightly increase. This is attributable to the anticipation of higher jumps and greater nonlinearities and complexities in the deformation profiles. \textcolor{black}{Also, note that the morphology of unit cell 2 presents more interfaces between both phases than unit cell 1, which can be one reason for higher errors in this test case. Nevertheless, we emphasize that the obtained results also depend on the shape and topology of the limited random samples chosen for the training. Therefore, one cannot simply conclude any relation between the topology of the microstructure and the accumulated error.} 

\color{black}

\subsection{Discussions on up-scaling results and mesh convergence}
\begin{figure}[H] 
  \centering
  \includegraphics[width=0.95\linewidth]{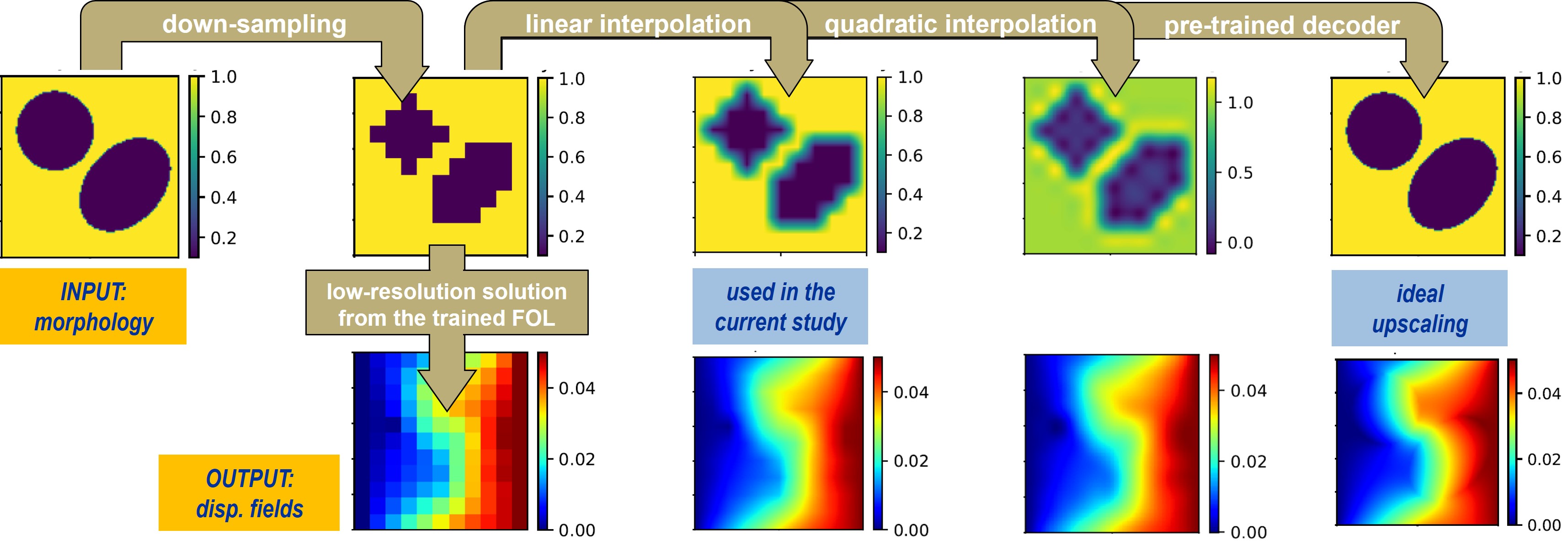}
  \caption{ Different strategies for upsampling the results (displacement component $U$). }
  \label{fig:upsampling}
\end{figure}

To minimize training costs, the microstructure morphologies (i.e., Young modulus distributions) undergo downsampling onto a smaller grid. The question arises regarding potential information loss and methods to reconstruct the original solution. In Fig.\,\ref{fig:upsampling}, various approaches are demonstrated for this purpose. The first method involves linear interpolation of the solution in the domain. Quadratic interpolation is also explored, but despite its smoothness, it does not provide significant advantages. Certainly, the most effective yet more costly strategy involves utilizing a pre-trained autoencoder, as shown on the right-hand side of Fig.\ref{fig:upsampling}. 

\begin{figure}[H] 
  \centering
  \includegraphics[width=0.95\linewidth]{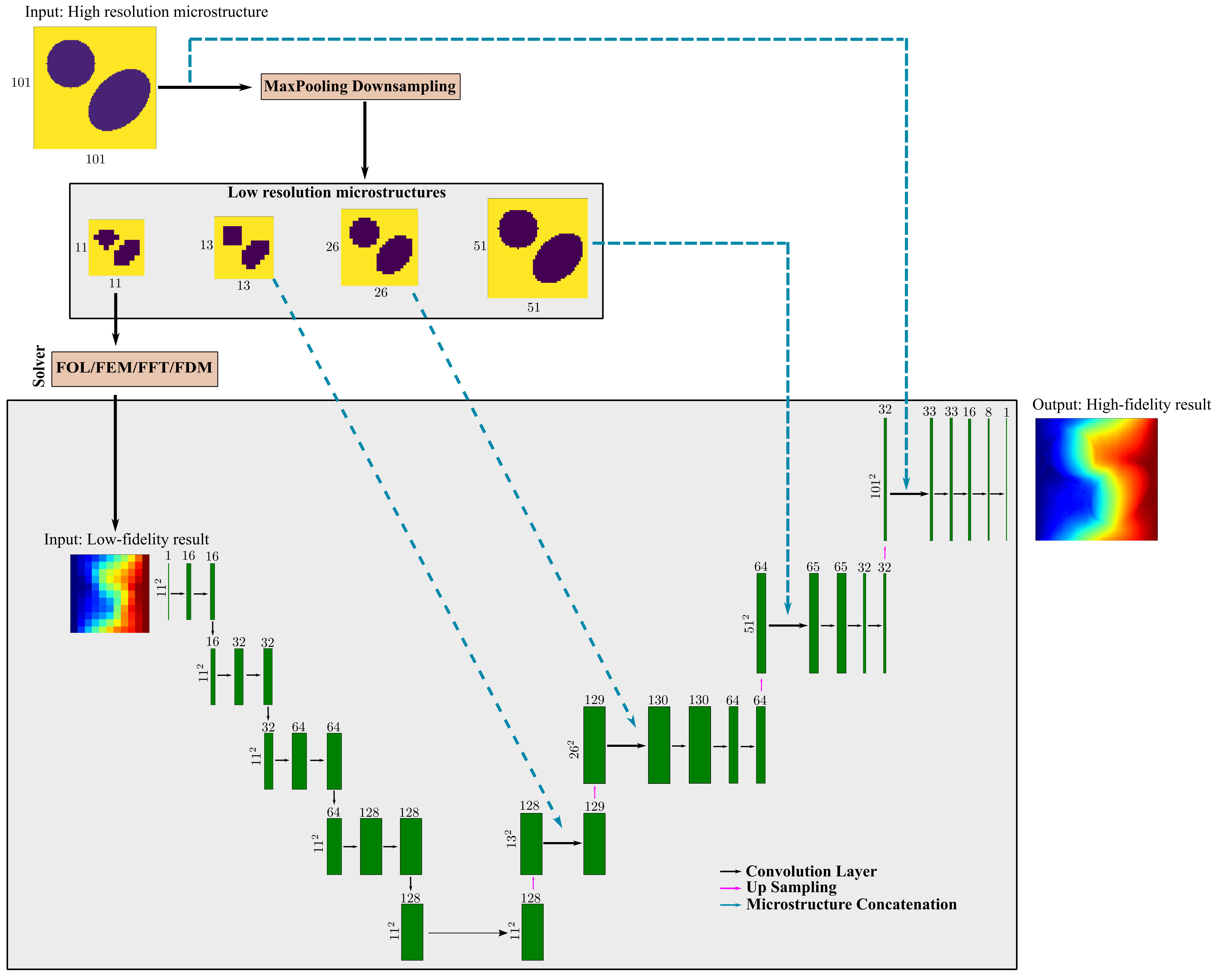}
  \caption{ In the developed MEA architecture, the process begins by condensing a high-resolution mechanical property map into various lower-resolution maps. The next step involves using the coarsest grid to solve the boundary value problem. The resultant low-resolution output is upscaled through an enhanced autoencoder.}
  \label{MAE_Mech}
\end{figure}

We report the results of a pre-trained model, referred to as the microstructure-embedded autoencoder (MEA). The original idea and similar works are introduced and reviewed in \cite{koopas2024introducing} and is summarized in Fig.\ref{MAE_Mech}. Firstly, the MEA method condenses a high-resolution parametric space, which represents the spatial distribution of mechanical properties for two-phase composite materials, into low-resolution grids. Secondly, it employs a pre-trained FOL to solve the boundary value problem on the coarsest grid. Finally, the solution is refined using the MEA architecture to produce a high-fidelity output by concatenation of parametric space fields of various resolutions in the decoder section.
The training of such an autodecoder requires solution data of the parameter space in different resolutions. The training of MEA with 5400 data sets takes 30 minutes on a workstation working with a Quadro RTX 8000 GPU.

\begin{figure}[H] 
  \centering
  \includegraphics[width=0.99\linewidth]{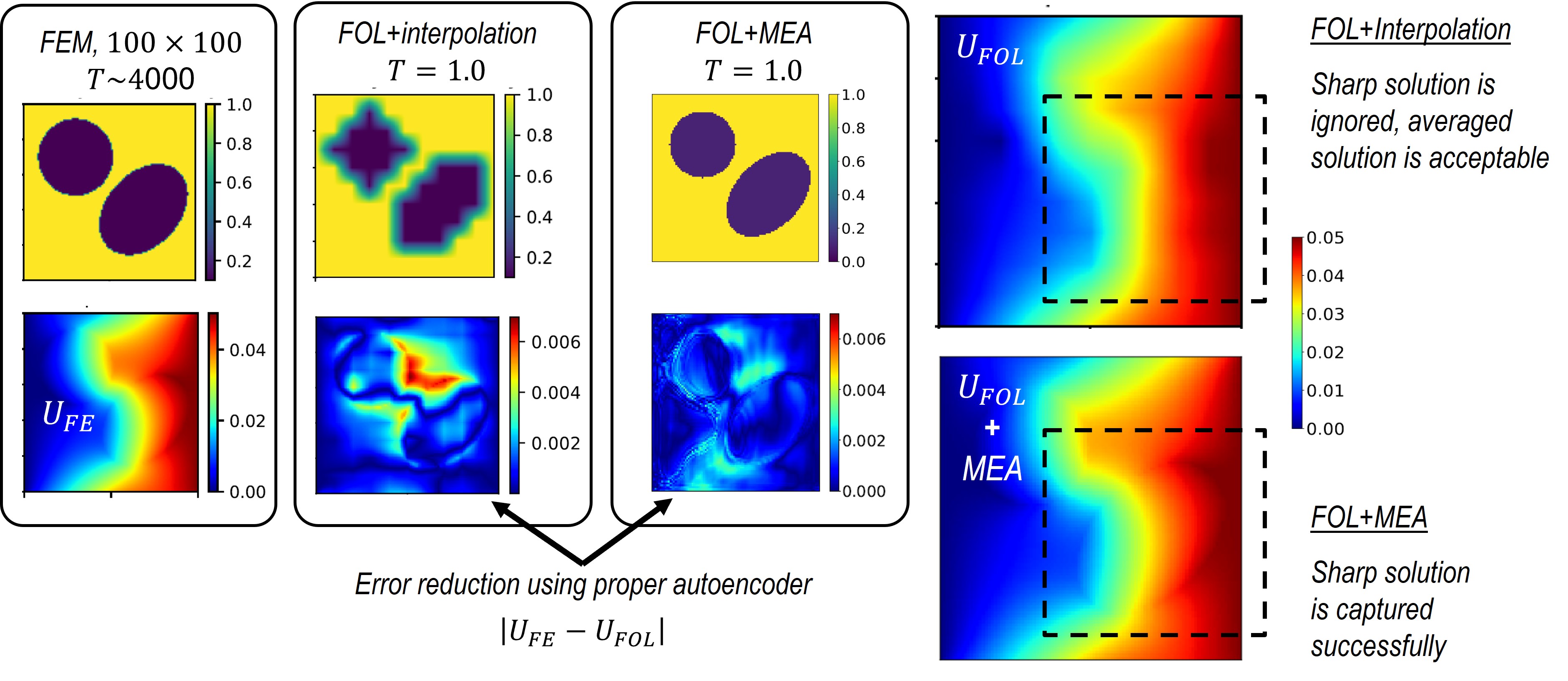}
  \caption{ Obtained results for the displacement component $U$ using the MEA approach show that errors are significantly reduced on a finer grid while reducing the computational cost substantially. }
  \label{fig:MEA_compare}
\end{figure}

Fig.~\ref{fig:MEA_compare} shows a significant error reduction compared to the simple interpolation approach using the MEA method. Despite its simplicity, it is worth mentioning that the interpolation approach is both cost-effective and implementable, resulting in an acceptable agreement when only averaged values of upsampled solutions are required for specific applications, such as homogenization. However, the MEA approach can capture the sharp solution transitions, which are important for the evaluation of local stress fields.
See also Fig.~\ref{fig:MEA_res} in Appendix C for more test cases using the introdcued MEA approach.

As shown above any neural operator techniques can be solely trained on the reduced space and then the information can be reconstructed via an additional deep learning model. Nevertheless, in some applications, it is much easier to train the neural operator based on the reduced set of input features from the beginning to avoid the above two-step process. The latter idea is discussed in the following section.

\subsection{Fourier-based parameterization for FOL: Mesh convergence and extension to 3D}
A possible shortcoming of the original FOL architecture introduced in Fig.~\ref{fig:NN_idea} is its training cost for higher mesh densities, as the number of neurons in the input and output layers increases based on the number of D.O.F. of the model. The latter is also a limiting factor for other neural operator techniques, as all of them require being fed a sufficient number and variety of input samples. Similar to the idea of parametric space reduction by means of simple downsampling strategies discussed earlier in this work, we propose an alternative way to significantly simplify the FOL architecture. As a result, we enhance the capability of the method to handle higher mesh densities and even 3D problems without a significant increase in training time.

\begin{figure}[H] 
  \centering
  \includegraphics[width=0.99\linewidth]{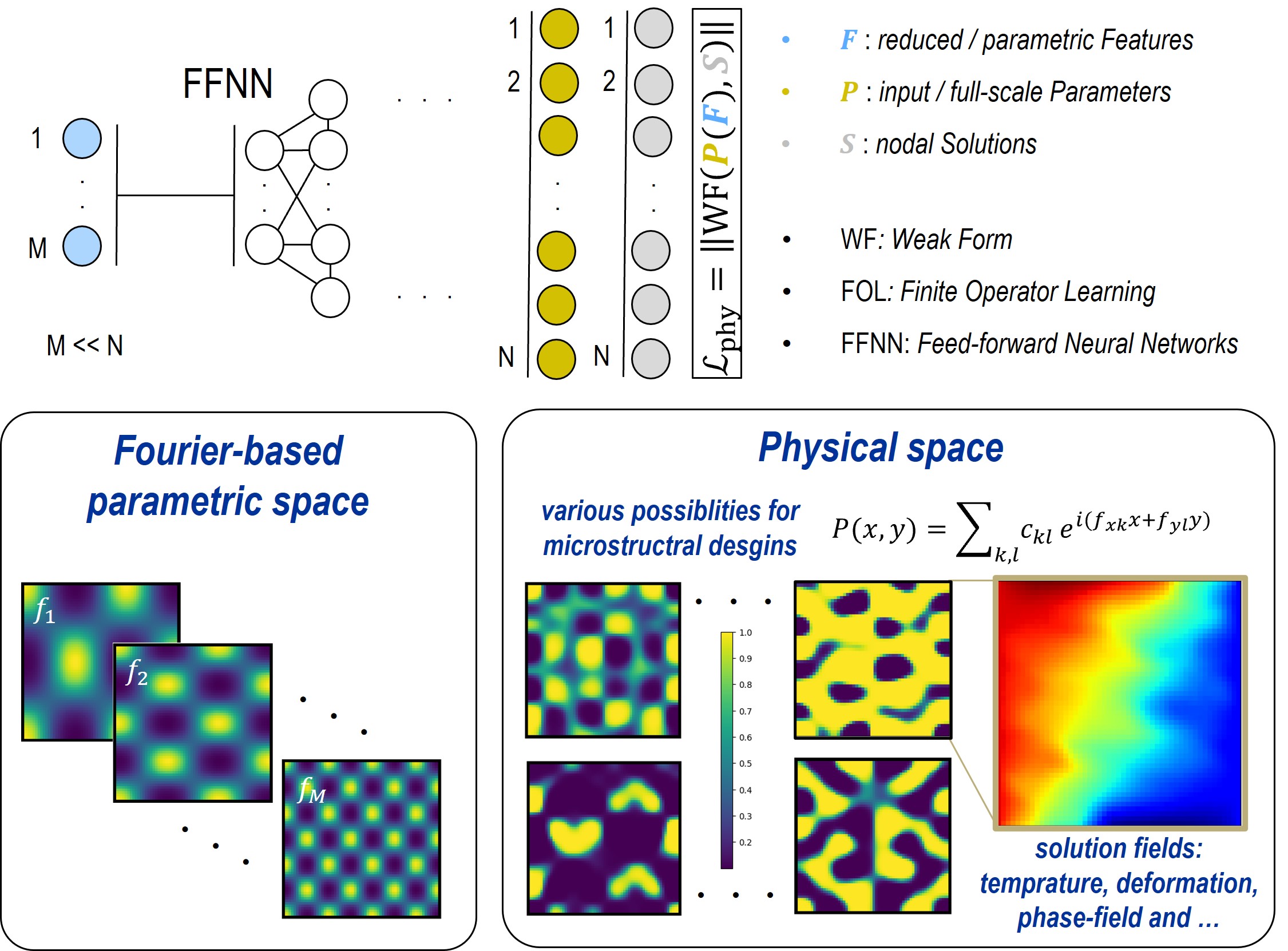}
  \caption{ The idea behind the Fourier-based parameterization for FOL is to decrease the input parametric space dimension and increase the resolution density of the output layer.}
  \label{fig:ff}
\end{figure}

This idea is based on the parametric representation of the inputs via a Fourier transformation where a combination of different frequencies for sinusoidal and cosinusoidal functions forms the input layer \cite{tancik2020fourier}. As depicted in Fig.\ref{fig:ff}, we can reduce the number of inputs to $M$ desired combinations of frequencies, which is usually a much smaller than the number of degrees of freedom in the system, denoted by $N$. Moreover, we take advantage of applying hard boundary constraints, and therefore only the first term in Eq.~\ref{eq:loss_sum} is used for minimization. We use $U_R = -0.05~$mm and $V_R = 0.1~$mm to show the flexibility of the method to handle other BCs. The application of hard boundaries in the FOL method is very straightforward, where we simply omit those degrees of freedom with Dirichlet boundary terms from the system of equations. This approach and similar ones are based on ideas from the implementation of the finite element method.
In other words, we do not need any extra modification of the values in the output layer, similar to other studies on the application of hard boundaries in the PINN framework (see \cite{Harandi2023} and references therein).


The Fourier-based function for representation of the elasticity map in a 2D setting has the following form  through a combination of sine and cosine functions:
\begin{equation}
\label{eq:foruier_eq}
\begin{aligned}
    E_f(x,y) &= \sum_i^{n_{sum}} [c_i + A_i \sin{(f_{x,i}~x)} \cos{(f_{y,i}~y)} + B_i \cos{(f_{x,i}~x)} \sin{(f_{y,i}~y)} \\ & + C_i\sin{(f_{x,i}~x)} \sin{(f_{y,i}~y)} + D_i \cos{(f_{x,i}~x)} \cos({f_{y,i}~y)}].
\end{aligned}
\end{equation}
In Eq.\,\ref{eq:foruier_eq}, $\bm x = (x, y)^T$, $c_i$ is the real-valued constant, $\{A_i,~B_i,~C_i,~D_i\}$ represents the amplitudes of the corresponding frequency number and $\{f_{x,i},~f_{y,i} \}$ represents the frequency in the $x$ and $y$-direction. 
Finally, we specify the property function $E_f$ through a so-called Sigmoidal projection as shown below to have more realistic and interesting characteristics for the microstructure.
\begin{equation}
\label{eq:sigmoid}
\begin{aligned}
    E(x,y) = (E_{max}-E_{min})\cdot\text{Sigmoid}\left(\beta(E_f-0.5)\right) + E_{min}. 
\end{aligned}
\end{equation}
With the above projection, we ensure that the value of Young's modulus lies between $E_{min}=0.1$ and $E_{max}=1.0$ MPa. With the $\beta$ parameter, one can also control the transition between the two phases, which is set to $\beta=1$ for the samples in this study. Some of the main modes, as well as possible samples that can be processed, are shown in the lower part of Fig.~\ref{fig:ff}.

The new design for the Fourier-based parameterization for FOL framework can therefore be summarized below:
\begin{align}
\label{eq:in_out}
    \bm{X} &= [A_i,~B_i,~C_i,~D_i],~~~\bm{Y} = [U_j,~V_j],~~~i= 1 \cdots M,~~~j = 1 \cdots N, \\
    \bm{Y} &= \mathcal{N}_{ff} (\bm{X}; \bm{\theta}),~~~\bm{\theta} = \{\bm{W},\bm{b}\}.
\end{align}
For the rest of the study, and for simplification purposes, only the constant term as well as the last term involving the multiplication of two cosine functions are considered (i.e., $A_i=0,~B_i=0,~C_i=0$). However, even by limiting ourselves to these terms, one can produce many designs for the property function $E_f$.
We select three frequencies for each direction. Considering the constant term, this choice gives us $M = 3 \times 3 + 1 = 10$ different terms, which can be added up to construct $E(x,y)$ according to Eq.~\ref{eq:sigmoid} and Eq.~\ref{eq:foruier_eq}.

\begin{figure}[H] 
  \centering
  \includegraphics[width=0.99\linewidth]{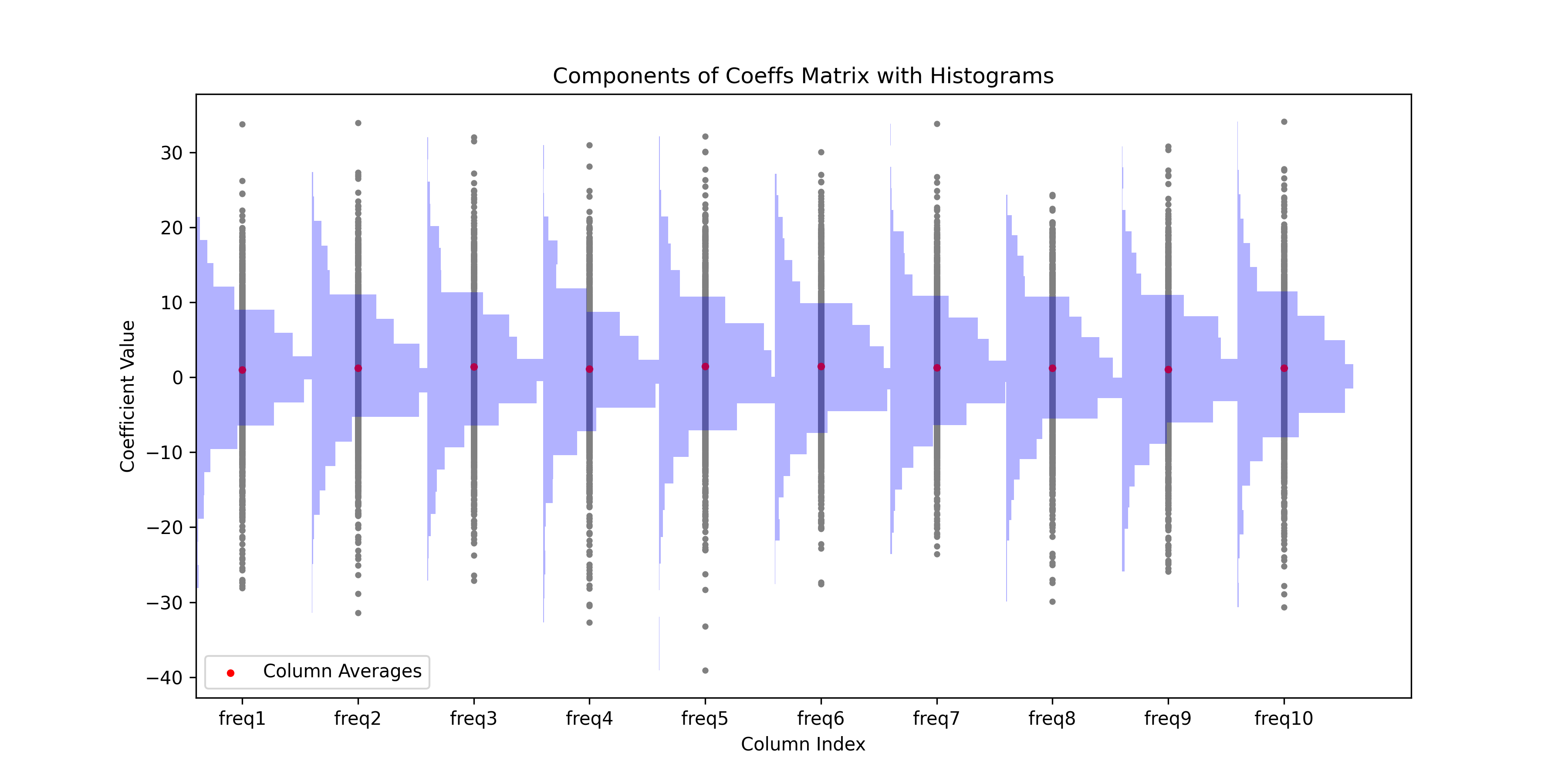}
  \caption{ Distribution of the 2000 random values for components of each frequency term. The parameter $\text{freq}_i$ represents the $i$-th coefficient corresponding to the $i$-th frequency. }
  \label{fig:coeffs_matrix}
\end{figure}

The input samples used for training are generated by combining different random distributions for these ten inputs. A summary of all the $2000$ selected samples is shown in Fig.~\ref{fig:coeffs_matrix}. See also Table \ref{Table:Fourier_param} for a summary of the Fourier-based parameterization for FOL model, as well as Fig.~\ref{fig:sample_2} for a few examples of samples obtained by the Fourier-based parameterization.

\begin{table}[H]
\centering
\caption{Parameters used for generating elasticity samples in training the Fourier-based FOL.}
 \begin{tabular}{ll} \\
    \hline
        Parameter &   Value\\
    \hline
     Frequency in the $x$- direction ${f_{x, i}}$ & $\left\{3,~5,~7\right\}$\\
     Frequency in the $y$- direction ${f_{y, i}}$ & $\left\{2,~4,~7\right\}$\\
     Number of random samples, batch size & $2000$, $1$ \\
     Neurons in the input layer (input features, $\text{freq}_i$) & $10$\\
     Mesh (Neurons in the output layer without DBCs) & $11 \times 11~(242)$, $21 \times 21~(882)$, $51 \times 51~(5202)$\\
     Act. func., hidden layers, learning rate, Epoch num. & Swish, [300, 300], 0.001, 5000 \\
    \hline\\    
    \end{tabular}
    \label{Table:Fourier_param}
\end{table}

Few unseen test cases and their corresponding Fourier coefficients are summarized in Table~\ref{Table:Fourier_coeff}. These cases are chosen in a way to examine if the proposed methodology can handle different possible scenarios. For example, in some cases, the matrix material is soft and reinforced by hard inclusions, while in others, the matrix is hard and includes soft parts (similar to voids).
It should be mentioned that for a given image of the microstructure, using the Fourier-based parameterization for FOL model, one must first perform an inverse Fourier analysis to find the corresponding coefficients for the input layer, which is a straightforward task.
\begin{table}[H]
  \centering
  \caption{Selected test cases and corresponding Fourier coefficients.}
   \begin{tabular}{lll} \\
      \hline
          Study type &   Value ($\text{freq}_i, i=1 \cdots 10$)\\
      \hline
       Mesh study, Fig.~\ref{fig:mesh_conv} & $\left[-4.0,2.0,-0.6,1.7,6.0,1.9,10.8,-3.6,-2.8,-9.9 \right]$\\
       Microstructure with voids, Figs.~\ref{fig:ff_10_u_m} and \ref{fig:ff_10_s_m}  & $\left[12.7, 6.1, -2.1, -14.1, -4.6, 2.1, 0.4, 2.7, 4.4, -3.1 \right]$\\
       Microstructure with inclusions, Fig.~\ref{fig:ff_12_u_m}  & $\left[-6.3, -0.4,  9.5,  4.4,  2.2,  0.9, 1.1,  0.4, -3.2, 0.3 \right]$\\
       Random Microstructure, Fig.~\ref{fig:ff_1929_u_m} & $\left[15.9, 17.2, 14.3,  2.0, 13.6,  2.5, -6.0, 7.4, -7.3, 8.2 \right]$\\
      \hline\\    
      \end{tabular}
      \label{Table:Fourier_coeff}
  \end{table}

\begin{figure}[H] 
  \centering
  \includegraphics[width=0.99\linewidth]{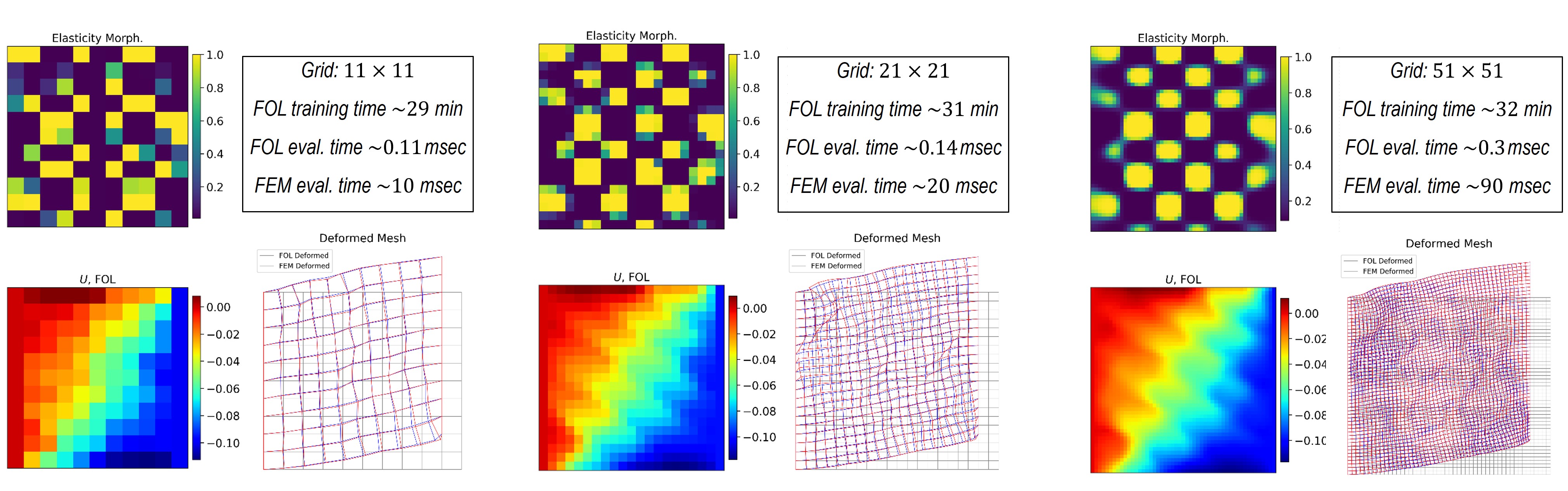}
  \caption{ Comparison of the results from FEM and Fourier-based parameterization for FOL. Convergence of the results with respect to the number of elements is demonstrated. The real-time computational cost using a Quadro RTX 6000 GPU for deep learning and CPU Ryzen pro7 for the FEM using FEAP software \cite{taylor_feap_2014}.  }
  \label{fig:mesh_conv}
\end{figure}

\begin{figure}[H] 
  \centering
  \includegraphics[width=0.80\linewidth]{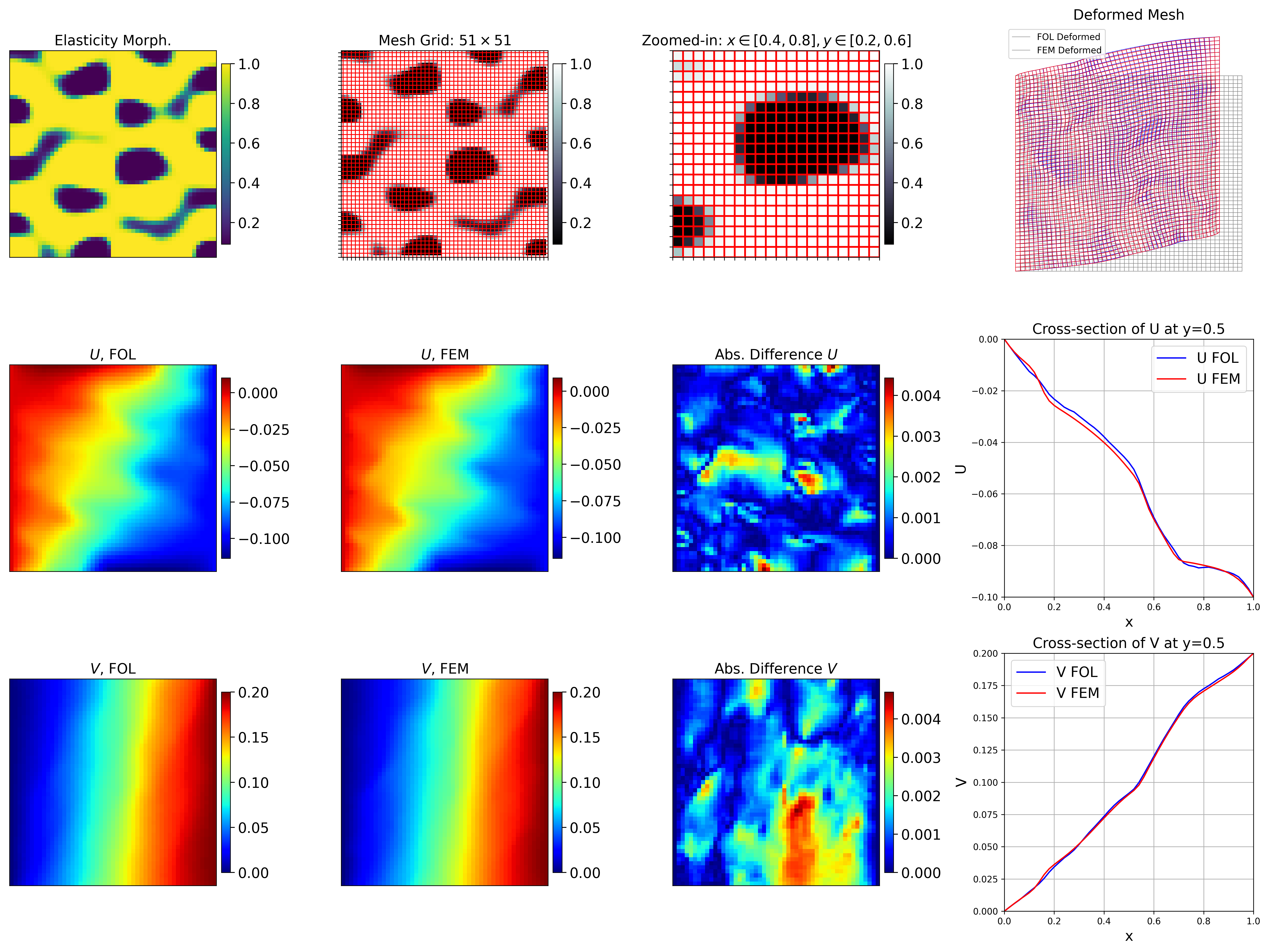}
  \caption{ Comparison of the deformation profile between FOL and FEM for a random microstructure using $\left[12.7, 6.1, -2.1, -14.1, -4.6, 2.1, 0.4, 2.7, 4.4, -3.1 \right]$. }
  \label{fig:ff_10_u_m}
\end{figure}

\begin{figure}[H] 
  \centering
  \includegraphics[width=0.80\linewidth]{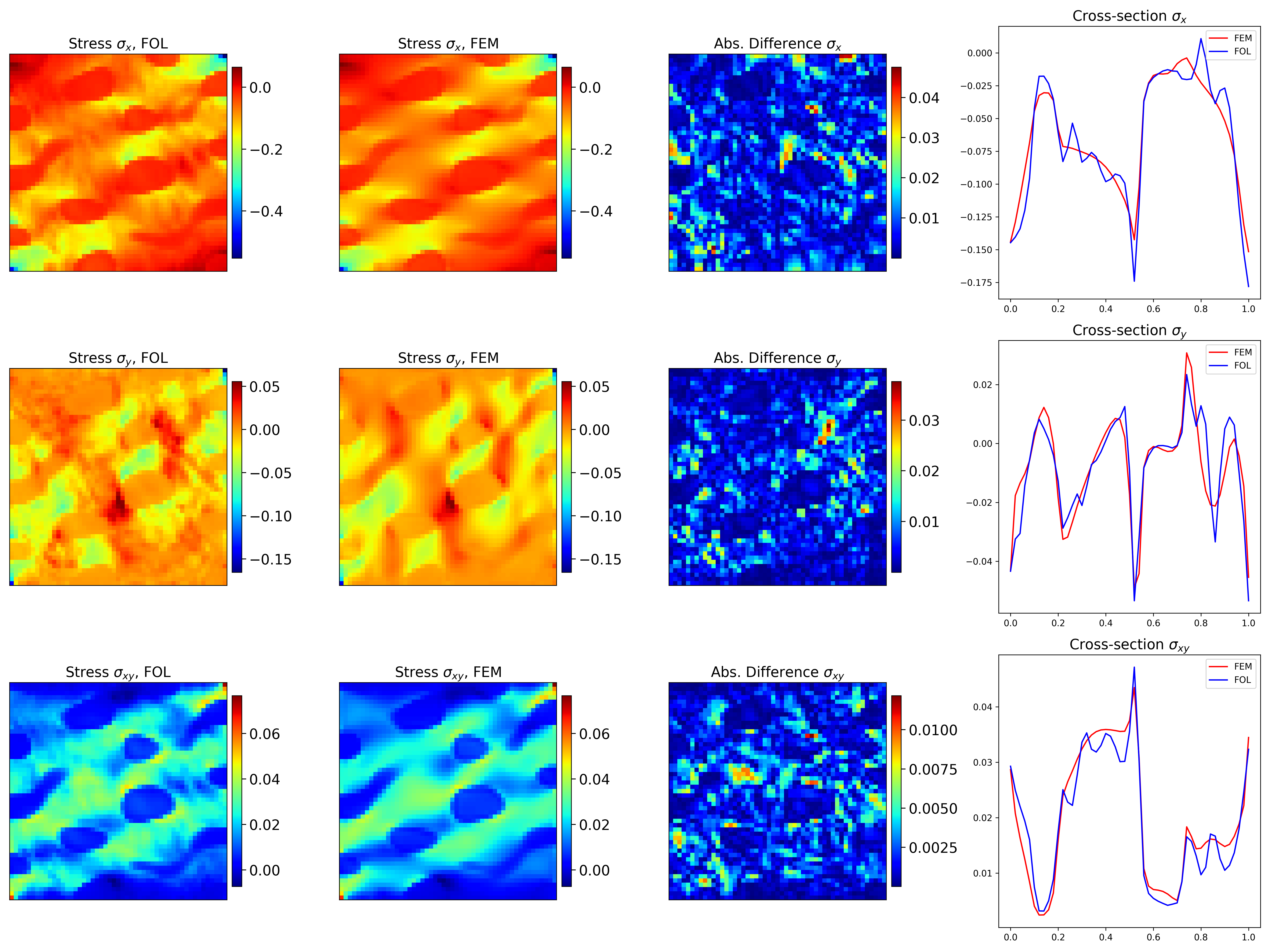}
  \caption{ Comparison of the stress profile between FOL and FEM for a random microstructure using $\left[12.7, 6.1, -2.1, -14.1, -4.6, 2.1, 0.4, 2.7, 4.4, -3.1 \right]$. }
  \label{fig:ff_10_s_m}
\end{figure}

\begin{figure}[H] 
  \centering
  \includegraphics[width=0.80\linewidth]{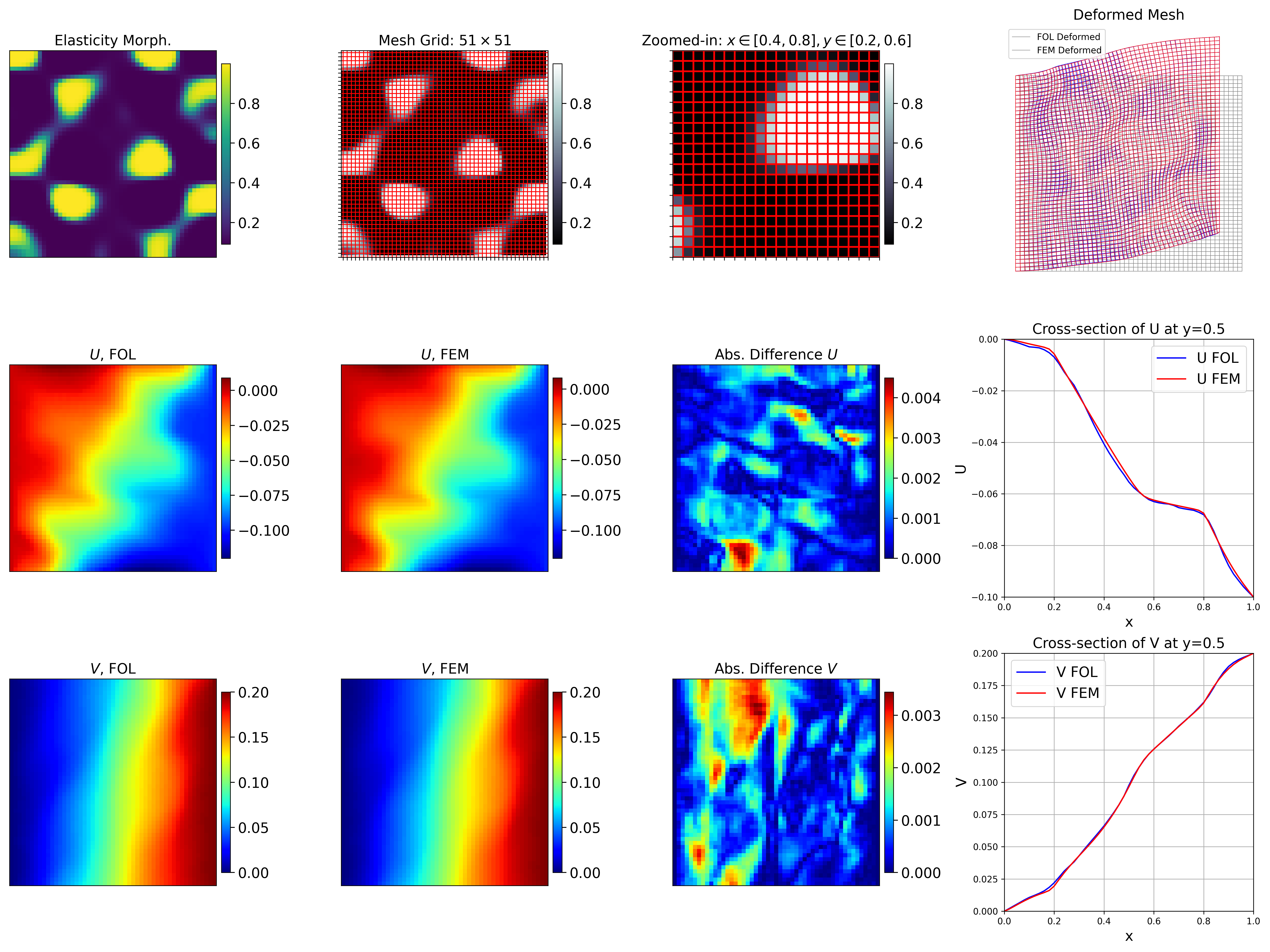}
  \caption{ Comparison of the deformation profile between FOL and FEM for a random microstructure using $\left[-6.3, -0.4,  9.5,  4.4,  2.2,  0.9, 1.1,  0.4, -3.2, 0.3 \right]$. }
  \label{fig:ff_12_u_m}
\end{figure}

\begin{figure}[H] 
  \centering
  \includegraphics[width=0.80\linewidth]{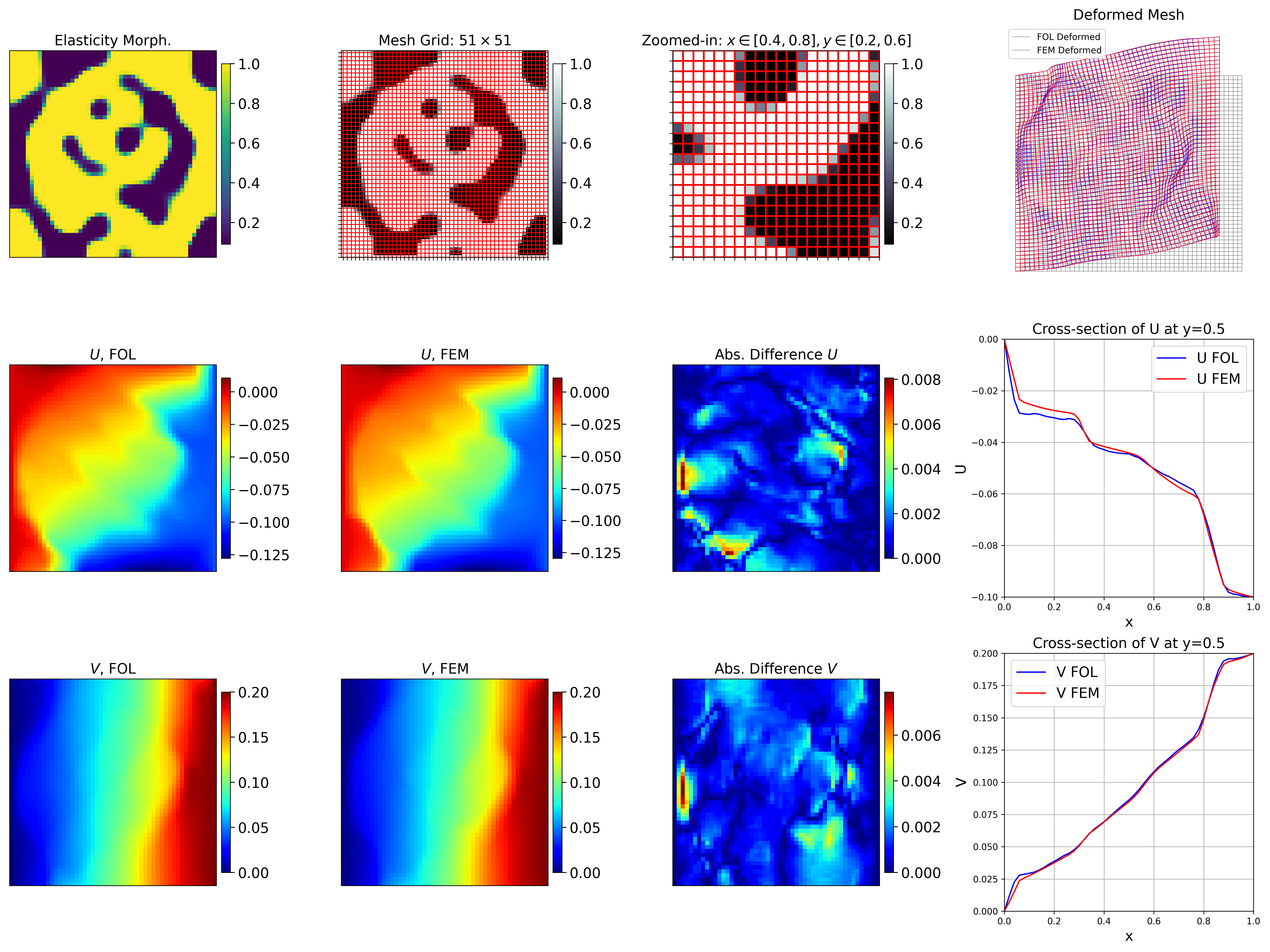}
  \caption{ Comparison of the deformation profile between FOL and FEM for a random microstructure using $\left[15.9, 17.2, 14.3,  2.0, 13.6,  2.5, -6.0, 7.4, -7.3, 8.2 \right]$. }
  \label{fig:ff_1929_u_m}
\end{figure}

Thanks to the new design of the FOL, we increase the final mesh resolution and avoid additional costs for training. We have checked this matter by testing the idea for different choices of $N$ as shown in Fig.~\ref{fig:mesh_conv}. Not only do the training and evaluation times (or costs) remain almost unchanged by dramatically increasing the neurons in the output layer from $121$ to $2061$, but we also observe a mesh convergence in the results.
All the training is performed using exactly the same hyperparameters for the deep learning model to ensure a fair comparison. However, based on the given resolution for the output layer, we can tune each neural network more effectively. The latter point is beyond the scope of the current study.
Moreover, we observe significant reduction in evaluation cost as we go to higher mesh densities. For the FEM calculations we do not claim that this implementation is the fastest available one.
The results reported in Figs.~\ref{fig:ff_10_u_m}, \ref{fig:ff_10_s_m}, \ref{fig:ff_12_u_m}, and \ref{fig:ff_1929_u_m} show an accepable agreement between the two methods. All these results are obtained for the $51 \times 51$ and no interpolation or any other additional post-processing step is done for the reported results. Very similar behavior is also observed for other mesh densities.

\newpage

\newpage

\subsection{Extension to 3D}
As mentioned before, FOL does not require any data and solely performs based on minimizing the discretized energy form or solving the corresponding discretized residual vector.
In other words, if we consider one sample as an input and set the batch size for the deep learning model to one, we end up with the exact formulation for FEM. 

The only difference is that we minimize the loss using the ADAM or L-BFGS algorithm (i.e. an iterative solvers) instead of the standard stiffness or tangent matrix in FEM (i.e. direct solvers). 
In this particular case, where we are not seeking parametric learning and solely looking for the solution of one specific boundary value problem, the FOL approach is similar to the PINN or, more accurately, the DEM approach. However, instead of using automatic differentiation, we use shape functions to calculate the derivatives. 
Even in this case, despite the slower convergence rate, the FOL approach offers two significant benefits. Firstly, for some nonlinear problems, this aspect might be interesting since one can bypass the linearization part of the residuals. This aspect remains to be demonstrated in future developments. Secondly, our investigations indicate that by avoiding the construction of the stiffness matrix, we can save a significant amount of memory. This memory efficiency allows us to achieve much higher mesh densities with the same hardware and implementation setup. 
For instance, with the aforementioned CPU and GPU capabilities, we were unable to perform FEM analysis using a $50 \times 50 \times 50$ mesh and FEAP software, while the FOL computation was completed successfully.

\begin{figure}[H] 
  \centering
  \includegraphics[width=0.99\linewidth]{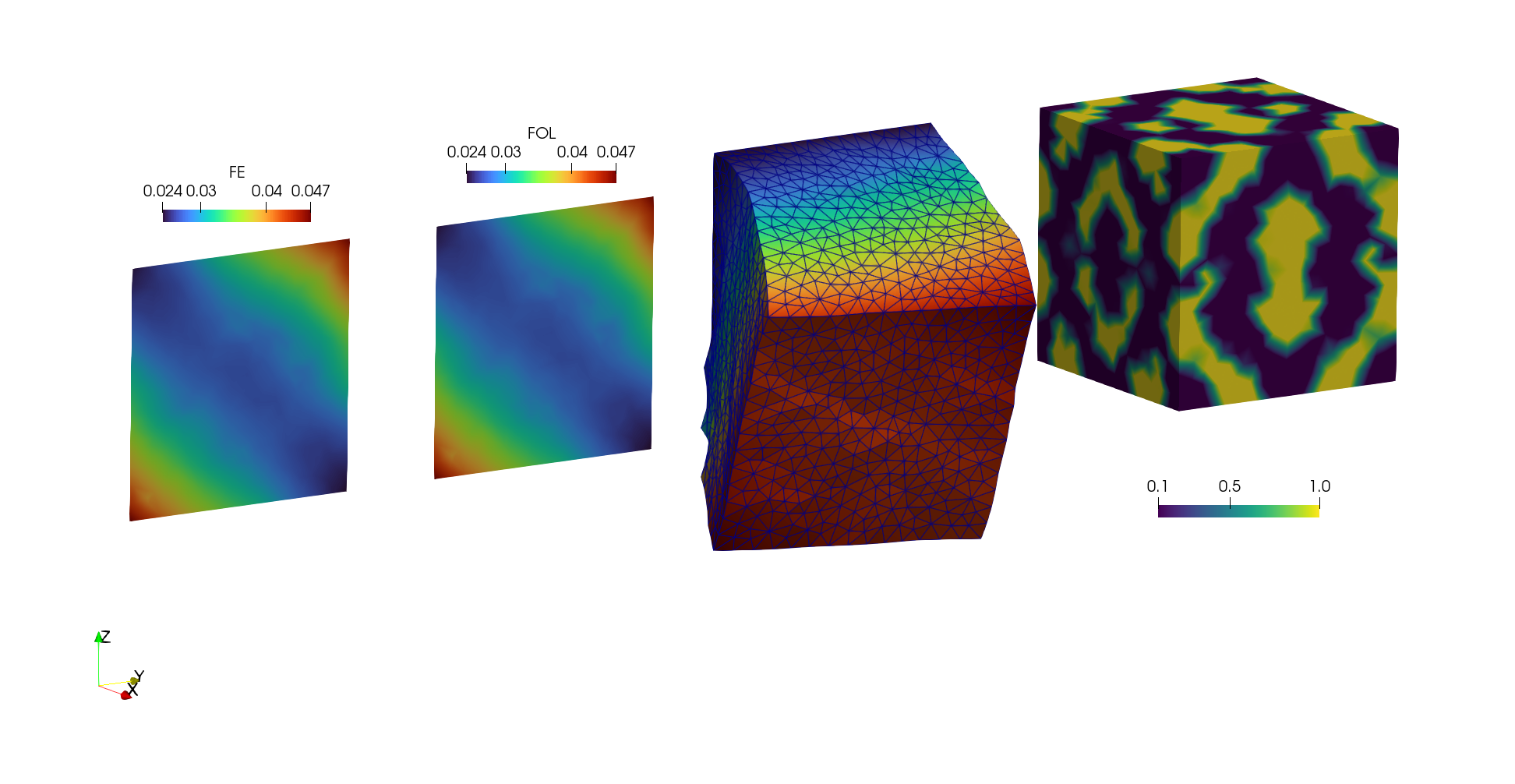}
  \caption{ Comparison of the results from FOL and FEM for a 3D example using unstructured tetrahedral elements. }
  \label{fig:final_3D_tetra}
\end{figure}

To show the applicability of the model for 3D problems, we investigated the following 3D example. 
The framework is again based on the Fourier-based FOL, and this time, three sets of frequencies are considered for the $x$, $y$, and $z$ directions. 
For the sample shown on the right hand side of Fig.~\ref{fig:final_3D_tetra}, we compare the results from the FEM and FOL in left side of Fig.~\ref{fig:final_3D_tetra}.
Note that in this study, we used unstructured tetrahedral elements for the discretization, which are known to be very flexible to enmesh complex geometries. For this example Gmsh \cite{Gmsh} is used.
Finally, the following Drichlet BCs are specified: the back surface fully fixed while the front surface is pulled via $\left[U_R=0.5,~V_R=-0.5,~W_R=0.0\right]$.

\begin{table}[H]
  \centering
  \caption{Parameters used for generating elasticity samples and training the Fourier-based FOL.}
   \begin{tabular}{ll} \\
      \hline
          Parameter &   Value\\
      \hline
       Frequency in the $x$- direction ${f_{x, i}}$ & $\left\{2,~4,~6\right\}$\\
       Frequency in the $y$- direction ${f_{y, i}}$ & $\left\{2,~4,~6\right\}$\\
       Frequency in the $z$- direction ${f_{z, i}}$ & $\left\{2,~4,~6\right\}$\\
       Number of random samples and batch size & $1$ \\
       Number of neurons in the input layer (input features, $\text{freq}_i$) & $28$\\
       Number of neurons in the output layer & $6920$ \\
       Activation func., hidden layers, learning rate, epoch & Swish, [1], 0.001, 500 \\
      \hline\\    
      \end{tabular}
      \label{Table:Fourier_param}
  \end{table}

\color{black}

\color{black}

\section{Conclusion and outlooks}
This work presents a data-free, physics-based method for parametric learning of partial differential equations in the field of mechanics of heterogeneous materials. This method relies solely on governing physical constraints, which are the discretized weak form of the problem as well as Dirichlet boundary terms. Leveraging the finite element method without depending on automatic differentiation for computing physical loss functions, the proposed approach approximates all derivatives using shape functions. This significantly improves training efficiency, necessitating fewer epochs and less time per epoch. Additionally, it allows flexibility in selecting activation functions without worrying about vanishing gradients. However, it's important to mention that potential discretization errors from the finite element method persist in the present approach.
\textcolor{black}{The introduced FOL framework can be seen as an extension of FEM and DEM, providing the ability to parametrically learn any interesting features (inputs) of the given boundary value problem. Moreover, the application of hard boundary conditions in the proposed framework is extremely straightforward, very similar to available methods in FEM.}

In this study, the network takes the elasticity distribution within the microstructure as input and produces the deformation components under fixed boundary conditions as output. Our method demonstrates accurate predictions of deformation fields, with the maximum pointwise error across all test cases consistently below $10\%$, primarily localized near phase boundaries. On the other hand, errors for first-order gradient terms, such as stress components, exhibit higher values. This observation requires further improvement in future developments, as local stress values significantly affect the localized nonlinear material behavior, such as plasticity and damage evolution. On the positive side, errors for averaged or homogenized values remain around $1\%$, making them acceptable for various engineering applications as long as the material is in the elastic regime. 
Despite possible errors and deviations in predicting the solution, the introduced methodology offers a significant speedup over the standard finite element method in elasticity. 
Furthermore, compared to a data-driven approach, the introduced physics-driven method demonstrates superior accuracy for unseen input parameters.
\textcolor{black}{We also discussed two different approaches to reduce the input parametric space while still achieving high-resolution solutions. 
The first approach is based on the microstructure-embedded autoencoder, which transfers the low-resolution solution to higher grid points through a pretrained autoencoder. The advantage of this approach is that any complex microstructure shape can be fed into the model to be reduced, solved, and upscaled. The potential drawback is that the training process of the autoencoder requires high-resolution FEM data. 
The second approach leverages Fourier-based parameterization for deep learning, where the input layer is significantly reduced to the coefficients of selected frequencies of sine and cosine functions in different spatial directions. The advantage of this approach is that we do not require any additional labeled data and can easily obtain the solution on much denser grids without any significant change in training time. However, obtaining the correct frequency coefficients for an arbitrary complex microstructure shape requires an additional step of an inverse Fourier transform. Both methods demonstrated excellent performance in predicting deformation patterns. The prediction of spatial derivatives, and therefore stress values, are also in good agreement, especially in the averaged sense. Ideas on how to improve them are discussed as well.}

The current investigations open up an entirely new approach for further research into combining physics-based operator learning and well-established numerical methods. Expanding this approach to nonlinear problems is essential, as it holds the potential for even greater speed improvements. While our focus was here on elastic problems in heterogeneous materials, exploring other types of equations, such as temporal problems and microstructure modeled by the phase field method presents an intriguing avenue for our future scope. Finally, it would be interesting to apply the current approach to inverse problems, wherein providing sufficient data can lead to discovering the underlying physics of the problem.
\\ \\ 
\textbf{Data Availability}:
The codes and data associated with this research are available upon request and will be published online following the official publication of the work.
\\ \\
\textbf{Acknowledgements}:
The authors would like to thank the Deutsche Forschungsgemeinschaft (DFG) for the funding support provided to develop the present work in the project Cluster of Excellence “Internet of Production” (project: 390621612). Ali Harandi and Stefanie Reese acknowledge the financial support of Transregional Collaborative Research Center SFB/TRR 339 with project number 453596084 funded by DFG gratefully. The author Rasoul Najafi Koopas would like to thank the Zentrum für Digitalisierungs- und Technologieforschung der Bundeswehr (dtec.bw) for their financial support.
\\ \\ 
\textbf{Author Statement}:
S.R.: Conceptualization, Methodology, Software, Writing - Review \& Editing. R.N.A.: Methodology, Software, Review \& Editing. S.F.: Supervision, Review \& Editing. M.A.: Software, Writing - Review \& Editing. A.H.: Software, Writing - Review \& Editing. R.N.K.: Software, Review \& Editing. G.L.: Supervision, Review \& Editing. S.R.: Supervision, Review \& Editing. M.A.: Funding, Supervision, Review \& Editing.

\newpage
\section{Appendix A: basics of the continuum theory and finite element discretization}

The kinematic relation defines the strain tensor $\bm{\varepsilon}$ in terms of the deformation vector $\bm{U}^T=[U,~ V]$ and reads:
\begin{align}
\label{kinematics}
\bm{\varepsilon}\,=\,\text{sym}(\text{grad}(\bm{U})) = \nabla^s\bm{U} = \dfrac{1}{2}\left(\nabla \bm{U} + \nabla \bm{U}^T \right).
\end{align} 
In the context of linear elasticity, we define the elastic energy of the solid as 
\begin{align}
\label{lin_energy}  
\psi_{lin} = \dfrac{1}{2}~\bm{\varepsilon}:\mathbb{C}(\bm{X}):\bm{\varepsilon},
\end{align}
where $\mathbb{C}$ is the fourth-order elasticity tensor. Through the constitutive relation, one relates the stress tensor to the strain tensor via 
\begin{align}
\label{materiallaw4}  
{\bm{\sigma}} &= {\mathbb{C}}(\bm{X})~{\bm{\varepsilon}}, \\
\mathbb{C}(\bm{X}) &= \Lambda(\bm{X})~\mathbf{I} \otimes \mathbf{I} + 2\mu(\bm{X})~\mathbb{I}^s.
\end{align}
Defining $\mathbf{I}$ as the second-order identity tensor and $\mathbb{I}^s$ as the symmetric fourth-order identity tensor, the above relation can also be written in the following form
\begin{align}
\label{materiallaw_Lamme}  
\boldsymbol{\sigma} = \Lambda(\bm{X}) \, \text{tr}(\boldsymbol{\varepsilon}) \, \mathbf{I} + 2\mu(\bm{X}) \, \boldsymbol{\varepsilon}.
\end{align}
Here, we have position-dependent Lame constants which can be written in terms of Young modulus $E$ and Poisson ratio $\nu$ as $\Lambda=E\nu/[(1-2\nu)(1+\nu)]$ and $\mu= E/[2(1+\nu)]$.
Here, the elastic properties are phase-dependent and vary throughout the microstructure. Finally, the mechanical equilibrium in the absence of body force, as well as the Dirichlet and Neumann boundary conditions, are written as:
\begin{align}
\label{Equilbrium}
\text{div}({\bm{\sigma}}) = \text{div}(\mathbb{C}(\bm{X})\nabla^s\bm{U})\ &= \bm{0}~~~~\text{in}~~~ \Omega\\
\label{BcsMech_d}
\bm{U} &= \bar{\bm{U}}~~~\text{on}~~\Gamma_D \\ 
\label{BcsMech_n}
\bm{\sigma} \cdot \bm{n} = \bm{t} &= \bar{\bm{t}}~~~~\text{on}~~\Gamma_N
\end{align} 
In the above relations, $\Omega$ and $\Gamma$ denote the material points in the body and on the boundary area, respectively. Moreover, the Dirichlet and Neumann boundary conditions are introduced in Eq.~\ref{BcsMech_d} and Eq.~\ref{BcsMech_n}, respectively.
Rewriting Eq.~\ref{materiallaw4} in the Voigt notation, we have $\hat{\bm{\sigma}}=\boldsymbol C(\bm{X})\hat{\bm{\varepsilon}}$. Considering the plane stress assumption in 2D, we write:
\begin{align}
\label{elasticityPlanestrain}
\boldsymbol C(\bm{X})\,=\, \dfrac{E(\bm{X})}{1-\nu^2(\bm{X})}\,\begin{bmatrix} 1 & \nu(\bm{X}) & 0 \\
\nu(\bm{X}) & 1 & 0 \\
0 & 0 & \dfrac{1-\nu(\bm{X})}{2}
\end{bmatrix}.
\end{align}
It is assumed that the material behavior remains isotropic in each phase. 

By introducing $\delta\bm{U}^T=[\delta U,~\delta V]$ as standard test functions and performing integration by parts, the weak form of the mechanical equilibrium problem reads:
\begin{align}
\label{Weakform}
\int_{\Omega} \delta{\bm{\hat{\varepsilon}}}^T\,\boldsymbol C(\bm{X})\,\hat{\bm{\varepsilon}}\,~dV\,-\, \int_{\Gamma_N} \delta{\bm{U}^T}\,\bar{\bm{t}}~dA=0. 
\end{align} 
The weak form in Eq.~\ref{Weakform} can also be interpreted as the balance between the mechanical internal energy $E^M_{\text{int}}$ and mechanical external energy $E^M_{\text{ext}}$. 

Next, we shortly summarize the corresponding linear shape functions $\bm{N}$ and the deformation matrix $\bm{B}$ used to discretize the mechanical weak form in the current work.
\begin{align}
\label{eq:N}
\bm{N} =
\begin{bmatrix}
N_1 & 0 & \dots & N_4 & 0\\
0 & N_1 & \dots & 0 & N_4
\end{bmatrix},~
\bm{B} =
\begin{bmatrix}
N_{1,x} & 0       & \dots & N_{4,x} & 0        \\
0       & N_{1,y} & \dots & 0       & N_{4,y}  \\
N_{1,y} & N_{1,x} & \dots & N_{4,y} & N_{4,x} 
\end{bmatrix}.
\end{align}
The notation $N_{i,x}$ and $N_{i,y}$ represent the derivatives of the shape function $N_i$ with respect to the coordinates $x$ and $y$, respectively. To compute these derivatives, we utilize the Jacobian matrix
\begin{align}
 \boldsymbol J = \partial \boldsymbol X / \partial \boldsymbol \xi = \begin{bmatrix} \frac{\partial x}{\partial \xi} & \frac{\partial y}{\partial \xi} \\ \frac{\partial x}{\partial \eta} & \frac{\partial y}{\partial \eta} \end{bmatrix}. 
\end{align}

Here, $\boldsymbol X = [x, y]$ and $\boldsymbol \xi = [\xi, \eta]$ represent the physical and parent coordinate systems, respectively. It is worth mentioning that this determinant remains constant for parallelogram-shaped elements, eliminating the need to evaluate this term at each integration point.
Readers are also encouraged to see the standard procedure in any finite element subroutine \cite{bathe, hughes}.
Finally, for the B matrix, we have 
\begin{align}
 \bm{B} = \bm{J}^{-1} \begin{bmatrix} \frac{\partial N_1}{\partial \xi} & 0 & \frac{\partial N_2}{\partial \xi} & 0 & \frac{\partial N_3}{\partial \xi} & 0 & \frac{\partial N_4}{\partial \xi} & 0 \\ 0 & \frac{\partial N_1}{\partial \eta} & 0 & \frac{\partial N_2}{\partial \eta} & 0 & \frac{\partial N_3}{\partial \eta} & 0 & \frac{\partial N_4}{\partial \eta} \\ \frac{\partial N_1}{\partial \eta} & \frac{\partial N_1}{\partial \xi} & \frac{\partial N_2}{\partial \eta} & \frac{\partial N_2}{\partial \xi} & \frac{\partial N_3}{\partial \eta} & \frac{\partial N_3}{\partial \xi} & \frac{\partial N_4}{\partial \eta} & \frac{\partial N_4}{\partial \xi} \end{bmatrix}. 
\end{align}

\begin{figure}[H] 
  \centering
  \includegraphics[width=0.99\linewidth]{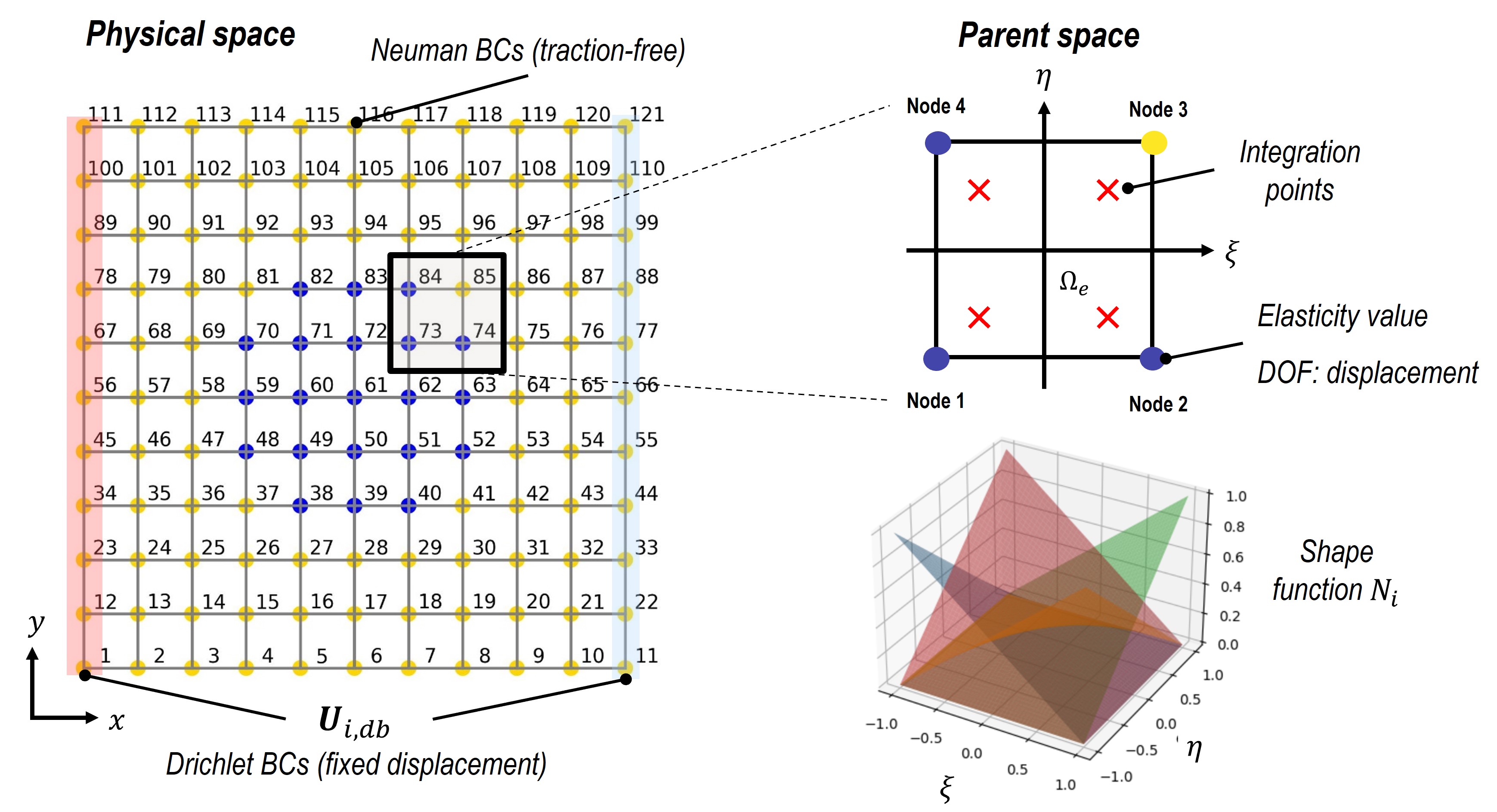}
  \caption{The physical space and the parent space. Gaussian integration and shape functions are shown specifically within the parent space. }
  \label{fig:FE_int}
\end{figure}
\color{black}

\newpage
\section{Appendix B: Details about the training of the DeepONet}
In this appendix, additional details about the training process of the DeepONet network are provided.
Sec.\,\ref{DeepONet} illustrates the vanilla DeepONet network architecture in Fig.\,
\ref{fig:DNOarch}.
The branch network processes the distribution of properties, which are measured at arbitrary sensor locations. Specifically, the input function to the branch network includes the values of Young's modulus at the element nodes. The branch maps the input function into two distinct vectors, one for each output. They can be considered as latent vectors, see \cite{boulle2023}. On the other hand, the trunk network is designed to receive the spatial coordinates of a single point where the output function is evaluated. Finally, two operators are built by the dot product of the last layers of the branch and the trunk. 

These operators are designed to map the input function $\bm{E}$, captured at $N$ sensor locations (nodes of elements), to the corresponding solutions at the given spatial coordinate $\bm{X}=(x_i, y_i)$ to the trunk network. 

In Eq.~\ref{eq:deeponetloss}, the loss function is defined as the mean squared error (MSE) between the predicted values of the operators for $U$ and $V$ and the solutions derived using the finite element method (FEM). For each input function, indexed by $i$ and covering the entire range of training inputs up to $I_{train}$, the total number of training input functions, the operator predicts the value at each point $j$. Then the difference between these predictions and the solutions obtained by FEM over $N$ points are computed. The network parameters, denoted as $\bm{\theta}$, are then optimized by minimizing the loss function outlined in Eq.~\ref{eq:deeponetloss}. 

For training the network, a purely data-driven approach is employed. Details regarding the sample preparation can be found in Sec.\,\ref{preparationofcollocationfields}. The training involves $4000$ samples, each featuring distinct morphology maps, and the training is done for $4000$ epochs. Since the architecture of the network differs significantly from the FOL approach, an extensive study of hyperparameters was performed to achieve optimal results. The best-performing configuration was found to be $6~[20]$, which indicates $6$ hidden layers when each layer consists of $20$ neurons. Notably, both the branch and the trunk networks use an identical architecture and share a common final layer containing $20$ neurons.
\begin{figure}[H] 
  \centering
  \includegraphics[width=0.50\linewidth]{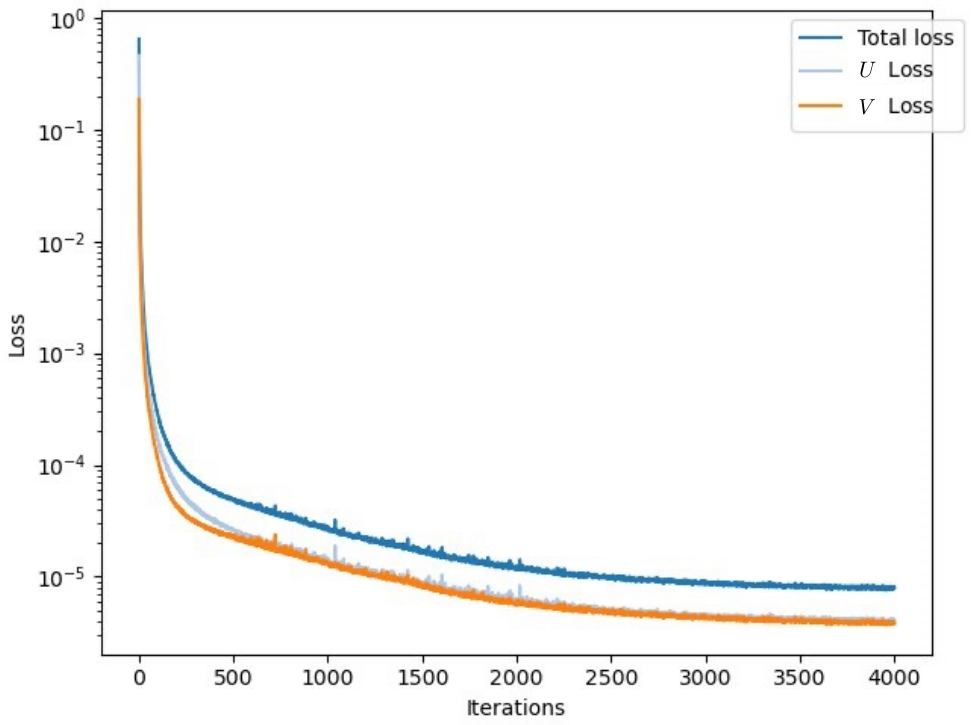}
  \caption{The decay of total loss and its $U$ and $V$ components for the DeepONet network with optimally found hyperparameters.}
  \label{fig:dno_loss}
\end{figure}
The best result is obtained by using the \emph{swish} as the activation function. The loss decay for the optimal hyperparameters is shown in Fig.\,\ref{fig:dno_loss}.

\color{black}

\newpage
\section{Appendix C: Results}

\begin{figure}[H] 
  \centering
  \includegraphics[width=1.0\linewidth]{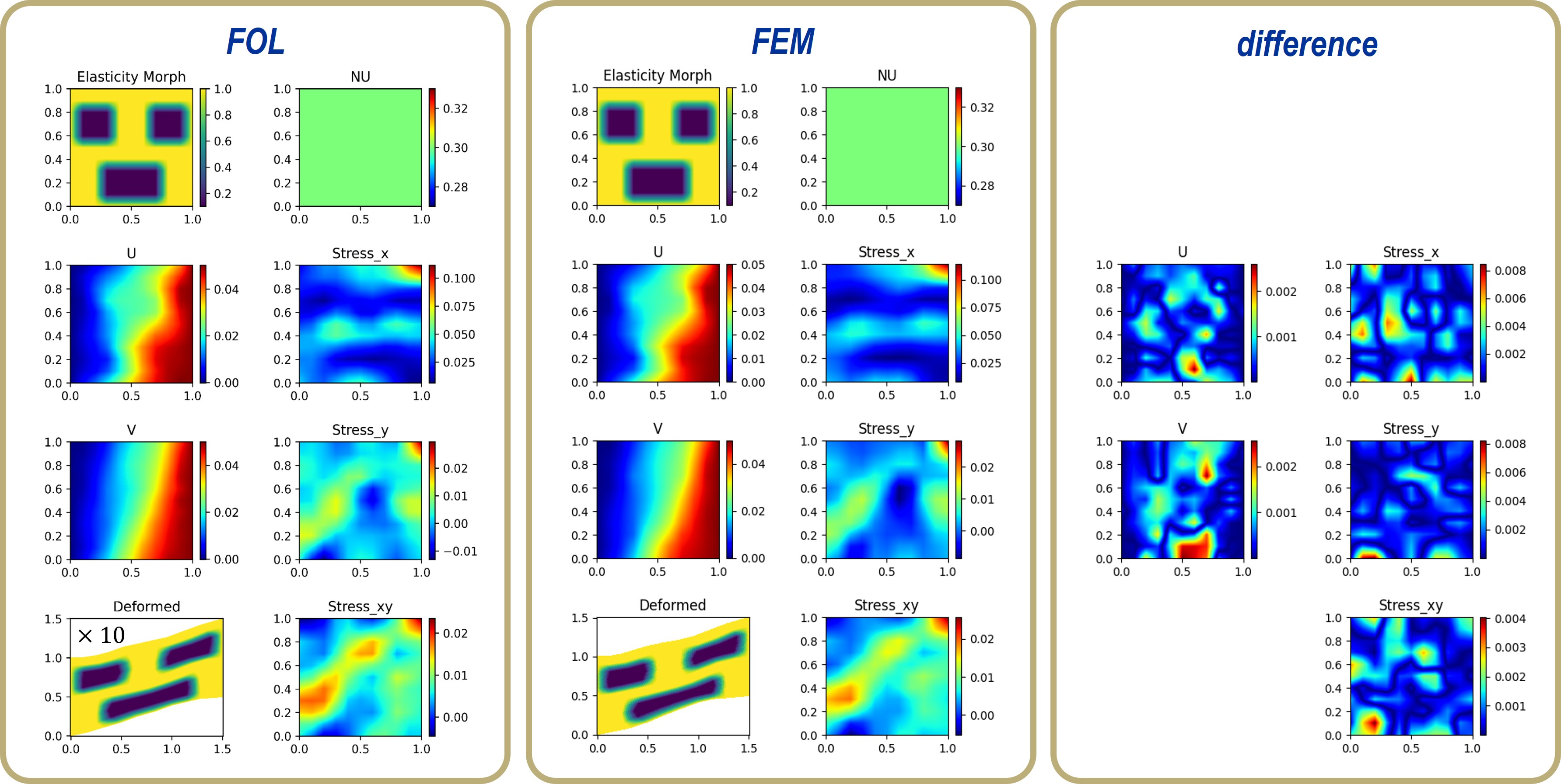}
  \caption{Predictions of the trained FOL for a test case with three square shape inclusions.}
  \label{fig:test0}
\end{figure}

\begin{figure}[H] 
  \centering
  \includegraphics[width=1.0\linewidth]{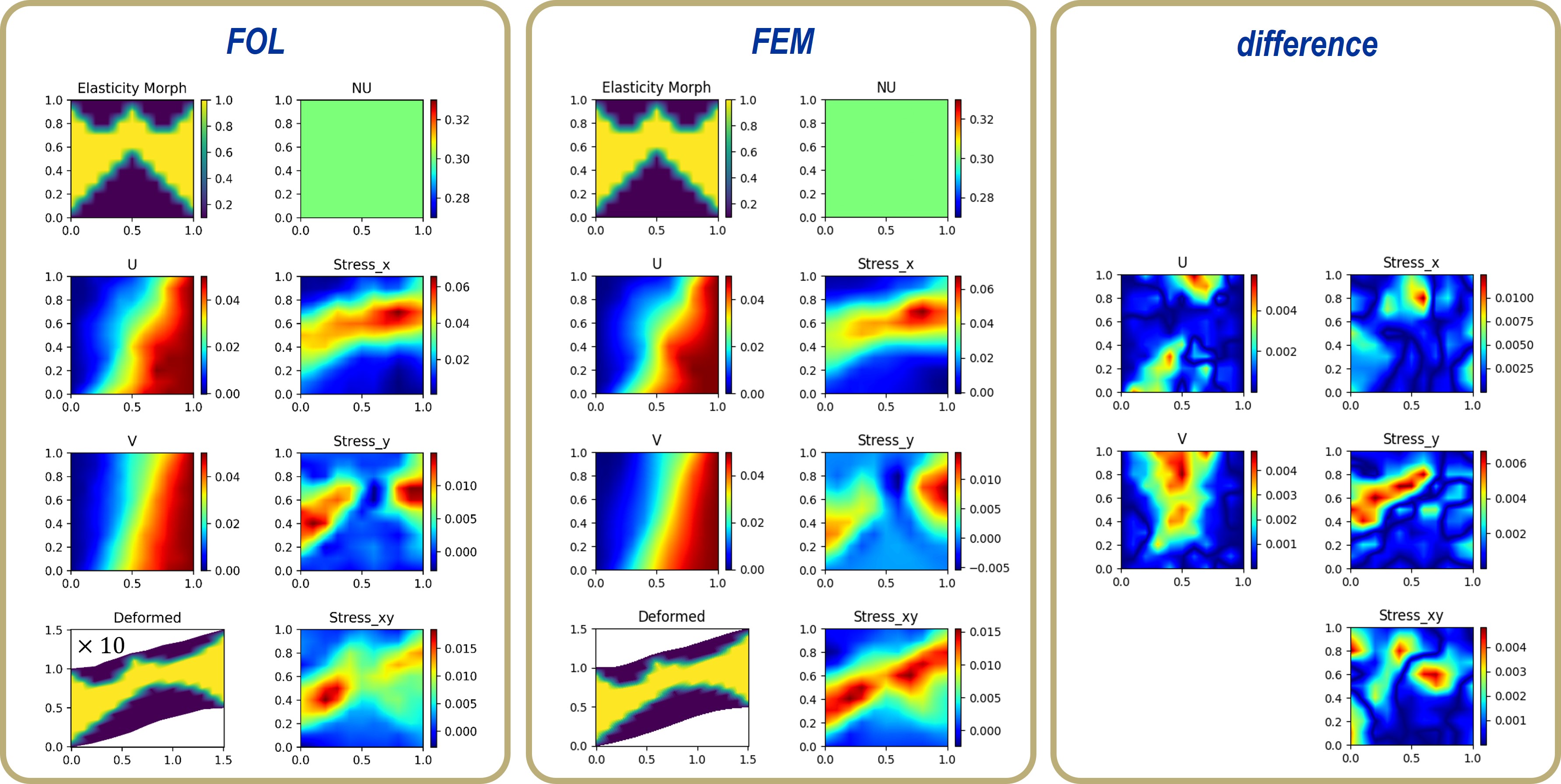}
  \caption{Predictions of the trained FOL for test similar to architectured materials which behave anisotropic.}
  \label{fig:test2}
\end{figure}

In Figs.~\ref{fig:test0} and ~\ref{fig:test2}, two other 2D microstructures are specified for the original FOL and their results are reported.

\begin{figure}[H] 
  \centering
  \includegraphics[width=0.99\linewidth]{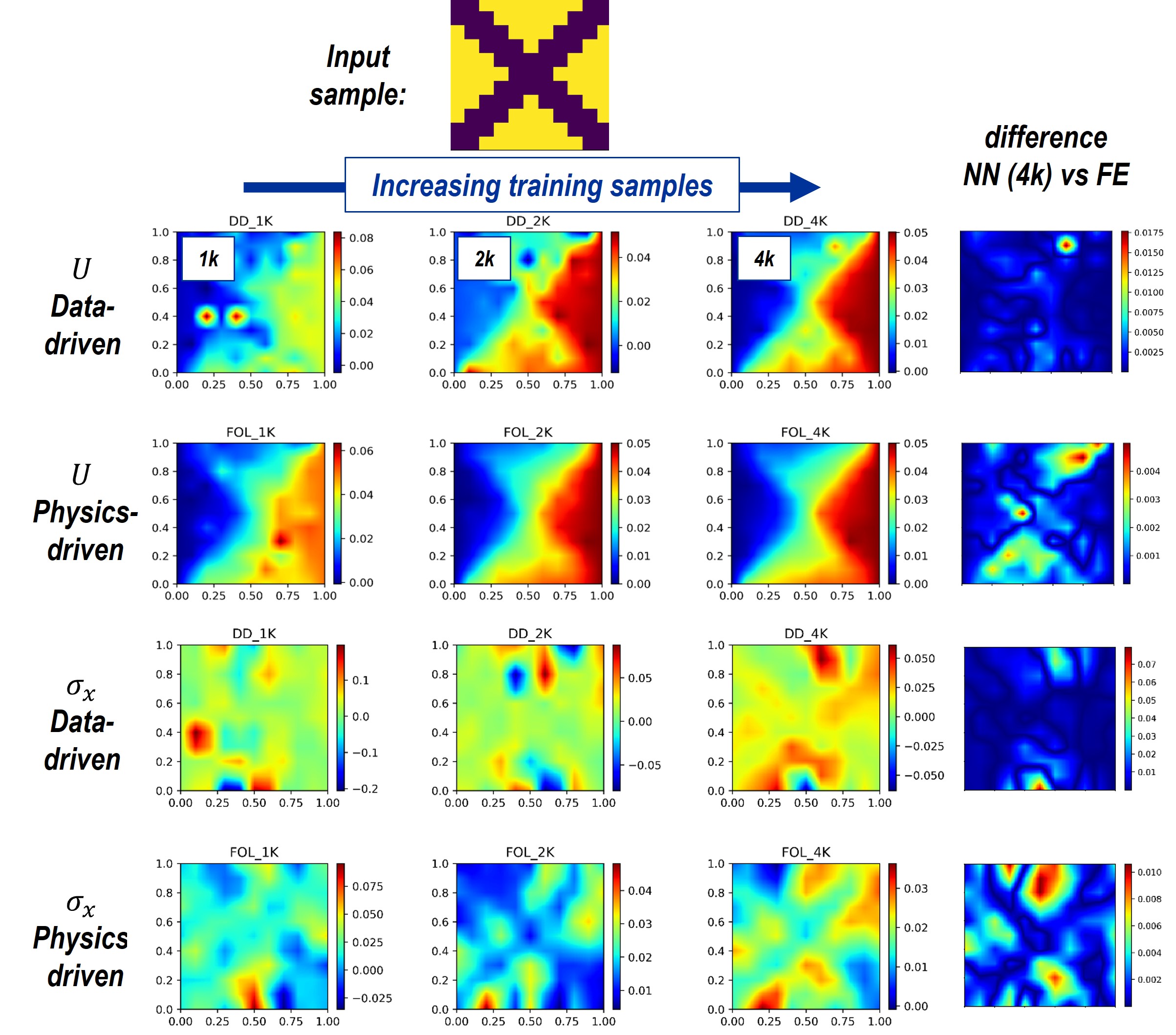}
  \caption{Performance of data-driven and physics-driven approaches for the different number of training samples.
   }
  \label{fig:FOL_conv_DD_2}
\end{figure}

\newpage
In Fig.~\ref{fig:FOL_conv_DD_2}, we show another test case to demonstrate the performance of the FOL framework with respect to the number of sample fields. Similar to the results reported in section 4.2, the predictions improve by feeding more samples to the training process while keeping all other parameters (including the NN settings) the same.

\begin{figure}[H] 
  \centering
  \includegraphics[width=0.95\linewidth]{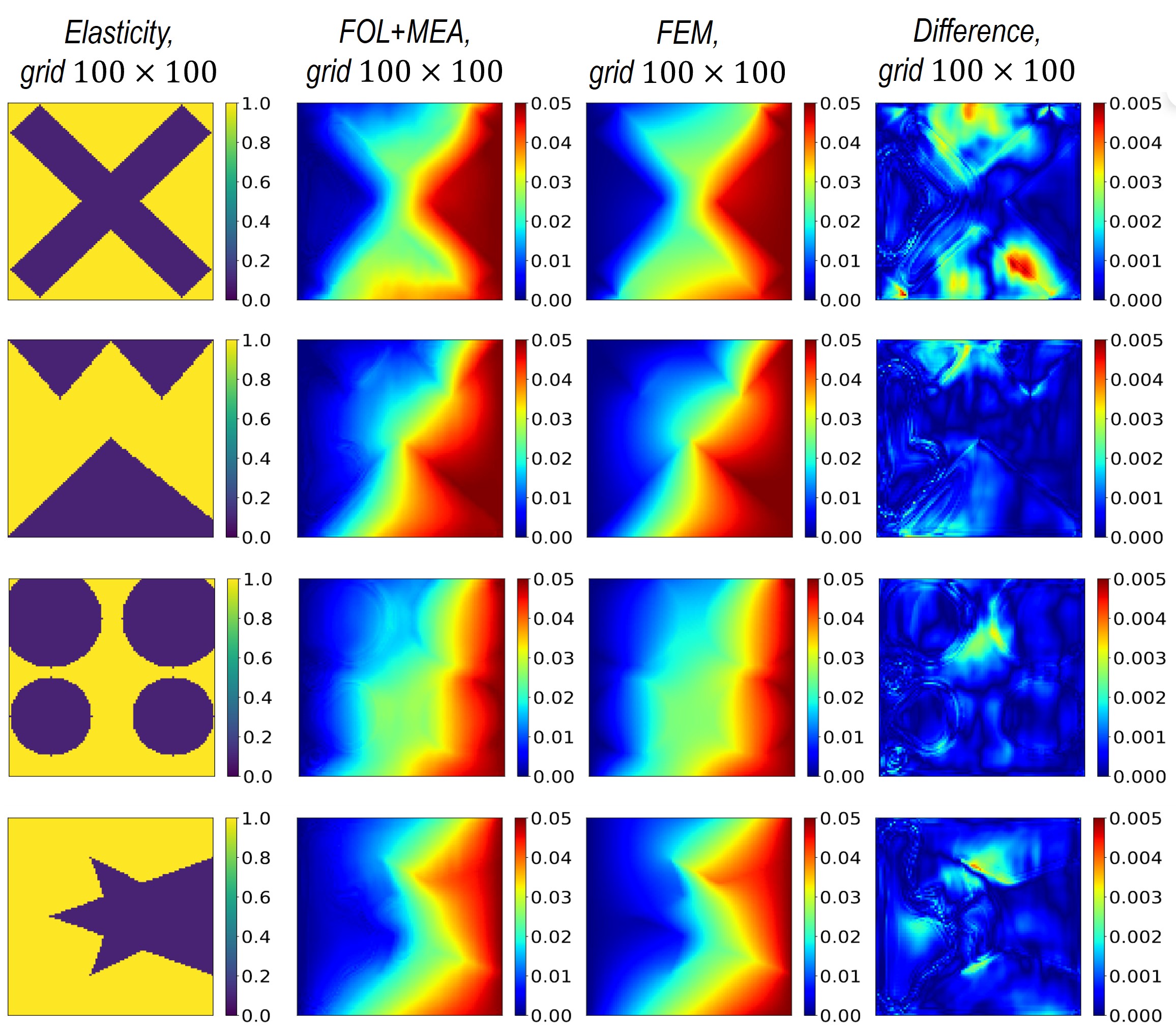}
  \caption{ Prediction of the microstructure-embedded autoencoder (MEA) method for the mechanical deformation on high resolution grid (displacement component $U$ evalauated on $101 \times 101$ grid). }
  \label{fig:MEA_res}
\end{figure}

\newpage
In Fig.~\ref{fig:MEA_res}, we provide more results on the performance of the MEA approach on different test cases. Note that none of these samples are used in the training of the FOL or in the training process of the MEA approach. As mentioned before, the MEA approach is able to map the low-resolution solution from the FOL (or even from any other solver, if applicable) onto a high-resolution grid space, capturing even sharp jumps in the solution quite effectively. Using the MEA approach will enable users to focus on training their neural operators solely based on the reduced topological space and input parameters, which makes the training efficient and more accurate.

\begin{figure}[H] 
  \centering
  \includegraphics[width=0.85\linewidth]{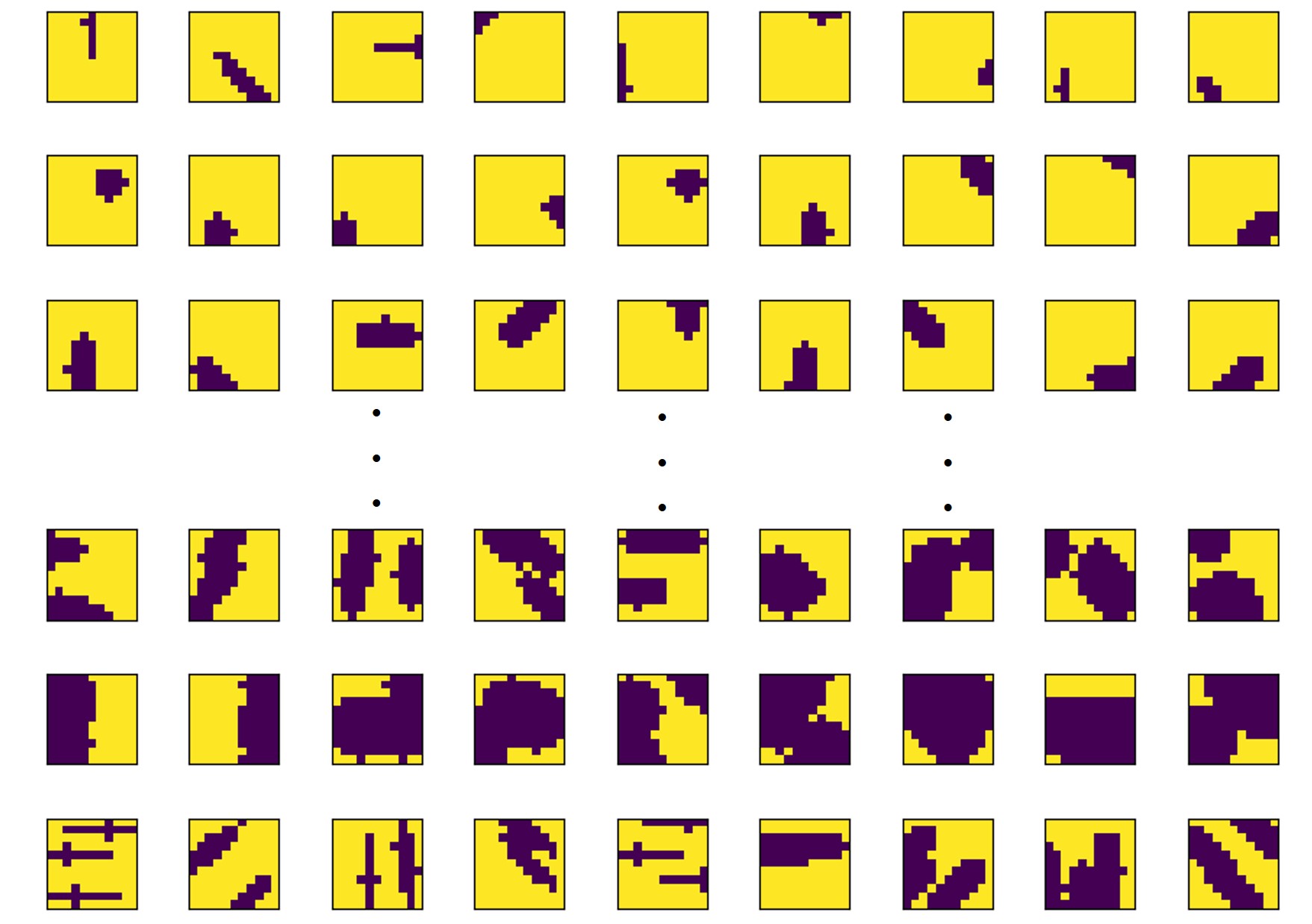}
  \caption{ Examples for random samples used for training of the NN described in section 3.3 and section 4.1.  The samples are shown on $11 \times 11$ grid which used for the traning of the FOL. Other resolutions of the samples are also available for training the MEA approach described in Section 4.5.}
  \label{fig:sample_1}
\end{figure}

\begin{figure}[H] 
  \centering
  \includegraphics[width=0.85\linewidth]{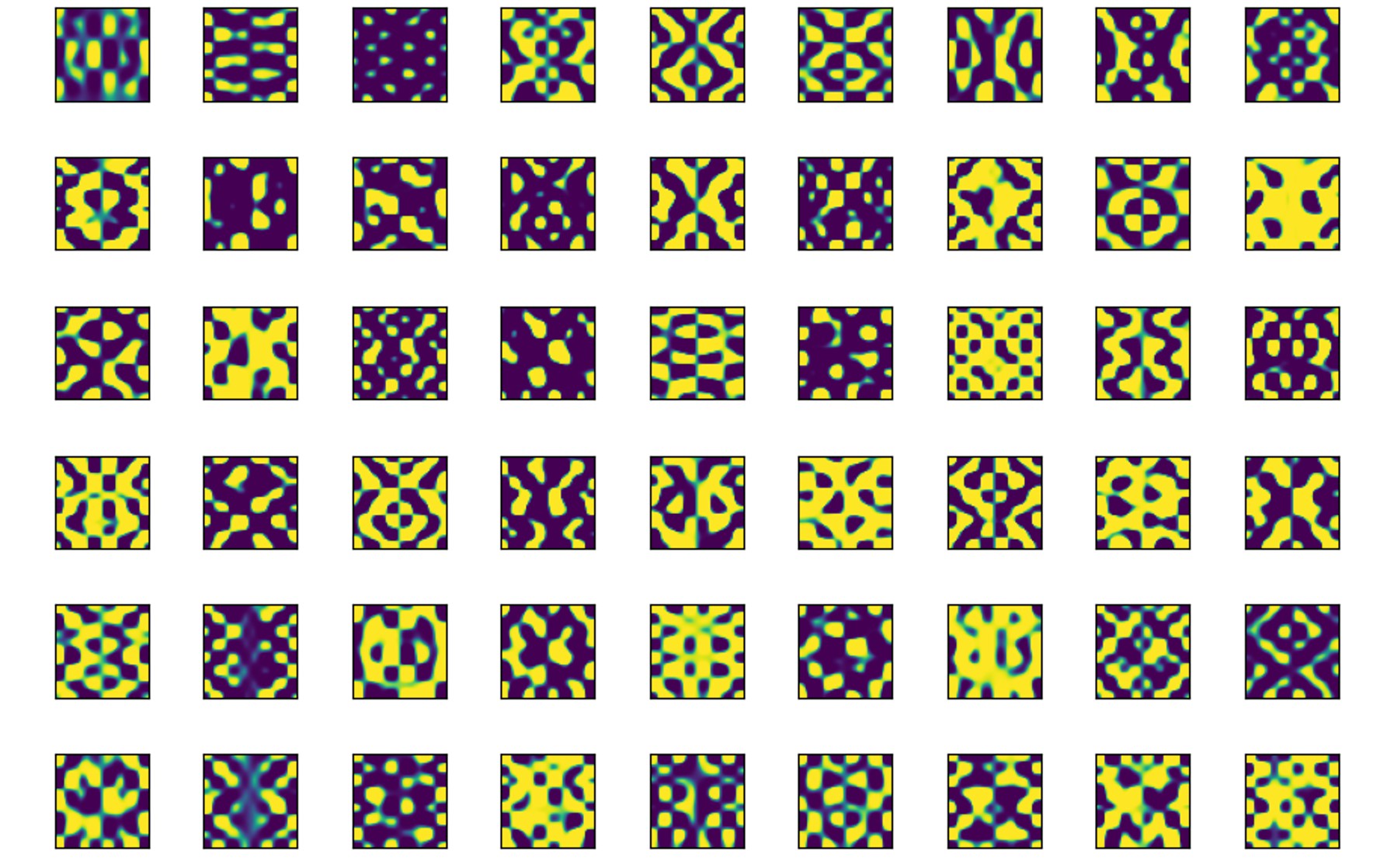}
  \caption{ Examples for random samples used for training of the NN described in section 4.6. The samples are constructed on $51 \times 51$ grid. }
  \label{fig:sample_2}
\end{figure}

In Figs.~\ref{fig:sample_1} and \ref{fig:sample_2}, few examples of the randomly generated samples for the training of the original FOL and Fourier-based FOL are shown, respectively. 

\color{black}

\newpage
\bibliography{Ref}

\begin{thebibliography}{62}
\providecommand{\natexlab}[1]{#1}
\providecommand{\url}[1]{\texttt{#1}}
\expandafter\ifx\csname urlstyle\endcsname\relax
  \providecommand{\doi}[1]{doi: #1}\else
  \providecommand{\doi}{doi: \begingroup \urlstyle{rm}\Url}\fi

\bibitem[Geers et~al.(2010)Geers, Kouznetsova, and Brekelmans]{GEERS20102175}
M.G.D. Geers, V.G. Kouznetsova, and W.A.M. Brekelmans.
\newblock Multi-scale computational homogenization: Trends and challenges.
\newblock \emph{Journal of Computational and Applied Mathematics}, 234\penalty0
  (7):\penalty0 2175--2182, 2010.

\bibitem[Faroughi et~al.(2024)Faroughi, Pawar, Fernandes, Raissi, Das,
  Kalantari, and Kourosh~Mahjour]{Faroughi2022}
Salah~A. Faroughi, Nikhil~M. Pawar, Célio Fernandes, Maziar Raissi, Subasish
  Das, Nima~K. Kalantari, and Seyed Kourosh~Mahjour.
\newblock {Physics-Guided, Physics-Informed, and Physics-Encoded Neural
  Networks and Operators in Scientific Computing: Fluid and Solid Mechanics}.
\newblock \emph{Journal of Computing and Information Science in Engineering},
  24\penalty0 (4):\penalty0 040802, 2024.

\bibitem[Liu et~al.(2022)Liu, Li, and Park]{Liu2022}
Wing~Kam Liu, Shaofan Li, and Harold~S. Park.
\newblock Eighty years of the finite element method: Birth, evolution, and
  future.
\newblock \emph{Archives of Computational Methods in Engineering}, 2022.

\bibitem[Herrmann and Kollmannsberger(2024)]{Herrmann2024}
Leon Herrmann and Stefan Kollmannsberger.
\newblock Deep learning in computational mechanics: a review.
\newblock \emph{Computational Mechanics}, 2024.

\bibitem[Kim and Lee(2024)]{Kim2024}
Dongjin Kim and Jaewook Lee.
\newblock A review of physics informed neural networks for multiscale analysis
  and inverse problems.
\newblock \emph{Communications Physics}, 2024.

\bibitem[Lu et~al.(2021)Lu, Jin, Pang, Zhang, and Karniadakis]{Lu2021}
Lu~Lu, Pengzhan Jin, Guofei Pang, Zhongqiang Zhang, and George Karniadakis.
\newblock Learning nonlinear operators via deeponet based on the universal
  approximation theorem of operators.
\newblock \emph{Nature Machine Intelligence}, 2021.

\bibitem[Li et~al.(2021)Li, Kovachki, Azizzadenesheli, Liu, Bhattacharya,
  Stuart, and Anandkumar]{li2021fourier}
Zongyi Li, Nikola Kovachki, Kamyar Azizzadenesheli, Burigede Liu, Kaushik
  Bhattacharya, Andrew Stuart, and Anima Anandkumar.
\newblock Fourier neural operator for parametric partial differential
  equations.
\newblock \emph{arXiv:2010.08895}, 2021.

\bibitem[Li et~al.(2020)Li, Kovachki, Azizzadenesheli, Liu, Bhattacharya,
  Stuart, and Anandkumar]{li2020neural}
Zongyi Li, Nikola Kovachki, Kamyar Azizzadenesheli, Burigede Liu, Kaushik
  Bhattacharya, Andrew Stuart, and Anima Anandkumar.
\newblock Neural operator: Graph kernel network for partial differential
  equations.
\newblock \emph{arXiv:2003.03485}, 2020.

\bibitem[Chen et~al.(2023)Chen, Liu, Meng, Chen, Liu, and Li]{chen2023learning}
Gengxiang Chen, Xu~Liu, Qinglu Meng, Lu~Chen, Changqing Liu, and Yingguang Li.
\newblock Learning neural operators on riemannian manifolds.
\newblock \emph{arXiv:2302.08166}, 2023.

\bibitem[Cao et~al.(2023)Cao, Goswami, and Karniadakis]{cao2023lno}
Qianying Cao, Somdatta Goswami, and George~Em Karniadakis.
\newblock Lno: Laplace neural operator for solving differential equations.
\newblock \emph{arXiv:2303.10528}, 2023.

\bibitem[Lu et~al.(2022)Lu, Meng, Cai, Mao, Goswami, Zhang, and
  Karniadakis]{LU2022114778}
Lu~Lu, Xuhui Meng, Shengze Cai, Zhiping Mao, Somdatta Goswami, Zhongqiang
  Zhang, and George~Em Karniadakis.
\newblock A comprehensive and fair comparison of two neural operators (with
  practical extensions) based on fair data.
\newblock \emph{Computer Methods in Applied Mechanics and Engineering},
  393:\penalty0 114778, 2022.

\bibitem[Rashid et~al.(2023)Rashid, Chakraborty, and
  Krishnan]{RASHID2023105444}
Meer~Mehran Rashid, Souvik Chakraborty, and N.M.~Anoop Krishnan.
\newblock Revealing the predictive power of neural operators for strain
  evolution in digital composites.
\newblock \emph{Journal of the Mechanics and Physics of Solids}, 181:\penalty0
  105444, 2023.

\bibitem[Lagaris et~al.(1998)Lagaris, Likas, and Fotiadis]{Lagaris1998}
I.E. Lagaris, A.~Likas, and D.I. Fotiadis.
\newblock Artificial neural networks for solving ordinary and partial
  differential equations.
\newblock \emph{IEEE Transactions on Neural Networks}, 9\penalty0 (5):\penalty0
  987--1000, 1998.

\bibitem[Sirignano and Spiliopoulos(2018)]{SIRIGNANO20181339}
Justin Sirignano and Konstantinos Spiliopoulos.
\newblock Dgm: A deep learning algorithm for solving partial differential
  equations.
\newblock \emph{Journal of Computational Physics}, 375:\penalty0 1339--1364,
  2018.

\bibitem[Raissi et~al.(2019)Raissi, Perdikaris, and Karniadakis]{RAISSI2019}
M.~Raissi, P.~Perdikaris, and G.E. Karniadakis.
\newblock Physics-informed neural networks: A deep learning framework for
  solving forward and inverse problems involving nonlinear partial differential
  equations.
\newblock \emph{Journal of Computational Physics}, 378:\penalty0 686--707,
  2019.

\bibitem[Rezaei et~al.(2022)Rezaei, Harandi, Moeineddin, Xu, and
  Reese]{REZAEI2022PINN}
Shahed Rezaei, Ali Harandi, Ahmad Moeineddin, Bai-Xiang Xu, and Stefanie Reese.
\newblock A mixed formulation for physics-informed neural networks as a
  potential solver for engineering problems in heterogeneous domains:
  Comparison with finite element method.
\newblock \emph{Computer Methods in Applied Mechanics and Engineering},
  401:\penalty0 115616, 2022.

\bibitem[Haghighat et~al.(2021)Haghighat, Raissi, Moure, Gomez, and
  Juanes]{HAGHIGHAT2021}
Ehsan Haghighat, Maziar Raissi, Adrian Moure, Hector Gomez, and Ruben Juanes.
\newblock A physics-informed deep learning framework for inversion and
  surrogate modeling in solid mechanics.
\newblock \emph{Computer Methods in Applied Mechanics and Engineering},
  379:\penalty0 113741, 2021.

\bibitem[Wu et~al.(2023)Wu, Jiang, Chen, Chatzigeorgiou, and
  Meraghni]{WU2023112521}
Jiajun Wu, Jindong Jiang, Qiang Chen, George Chatzigeorgiou, and Fodil
  Meraghni.
\newblock Deep homogenization networks for elastic heterogeneous materials with
  two- and three-dimensional periodicity.
\newblock \emph{International Journal of Solids and Structures}, 284:\penalty0
  112521, 2023.

\bibitem[Roy et~al.(2023)Roy, Bose, Sundararaghavan, and Arróyave]{ROY2023472}
Arunabha~M. Roy, Rikhi Bose, Veera Sundararaghavan, and Raymundo Arróyave.
\newblock Deep learning-accelerated computational framework based on physics
  informed neural network for the solution of linear elasticity.
\newblock \emph{Neural Networks}, 162:\penalty0 472--489, 2023.

\bibitem[Rezaei et~al.(2024{\natexlab{a}})Rezaei, Moeineddin, and
  Harandi]{rezaei2023learning}
Shahed Rezaei, Ahmad Moeineddin, and Ali Harandi.
\newblock Learning solutions of thermodynamics-based nonlinear constitutive
  material models using physics-informed neural networks.
\newblock \emph{Computational Mechanics}, 2024{\natexlab{a}}.

\bibitem[Haghighat et~al.(2023)Haghighat, Abouali, and
  Vaziri]{HAGHIGHAT2023105828}
Ehsan Haghighat, Sahar Abouali, and Reza Vaziri.
\newblock Constitutive model characterization and discovery using
  physics-informed deep learning.
\newblock \emph{Engineering Applications of Artificial Intelligence},
  120:\penalty0 105828, 2023.

\bibitem[Samaniego et~al.(2020)Samaniego, Anitescu, Goswami, Nguyen-Thanh, Guo,
  Hamdia, Zhuang, and Rabczuk]{SAMANIEGO2020112790}
E.~Samaniego, C.~Anitescu, S.~Goswami, V.M. Nguyen-Thanh, H.~Guo, K.~Hamdia,
  X.~Zhuang, and T.~Rabczuk.
\newblock An energy approach to the solution of partial differential equations
  in computational mechanics via machine learning: Concepts, implementation and
  applications.
\newblock \emph{Computer Methods in Applied Mechanics and Engineering},
  362:\penalty0 112790, 2020.

\bibitem[Nguyen-Thanh et~al.(2020)Nguyen-Thanh, Zhuang, and
  Rabczuk]{NGUYENTHANH2020103874}
Vien~Minh Nguyen-Thanh, Xiaoying Zhuang, and Timon Rabczuk.
\newblock A deep energy method for finite deformation hyperelasticity.
\newblock \emph{European Journal of Mechanics - A/Solids}, 80:\penalty0 103874,
  2020.

\bibitem[Fuhg and Bouklas(2022)]{FUHG2022110839}
Jan~N. Fuhg and Nikolaos Bouklas.
\newblock The mixed deep energy method for resolving concentration features in
  finite strain hyperelasticity.
\newblock \emph{Journal of Computational Physics}, 451:\penalty0 110839, 2022.

\bibitem[Jagtap et~al.(2020)Jagtap, Kharazmi, and
  Karniadakis]{JAGTAP2020113028}
Ameya~D. Jagtap, Ehsan Kharazmi, and George~Em Karniadakis.
\newblock Conservative physics-informed neural networks on discrete domains for
  conservation laws: Applications to forward and inverse problems.
\newblock \emph{Computer Methods in Applied Mechanics and Engineering},
  365:\penalty0 113028, 2020.

\bibitem[Kharazmi et~al.(2021)Kharazmi, Zhang, and
  Karniadakis]{KHARAZMI2021113547}
Ehsan Kharazmi, Zhongqiang Zhang, and George~E.M. Karniadakis.
\newblock hp-vpinns: Variational physics-informed neural networks with domain
  decomposition.
\newblock \emph{Computer Methods in Applied Mechanics and Engineering},
  374:\penalty0 113547, 2021.

\bibitem[McClenny and Braga-Neto(2020)]{McClenny22}
Levi McClenny and Ulisses Braga-Neto.
\newblock Self-adaptive physics-informed neural networks using a soft attention
  mechanism.
\newblock \emph{arXiv:2009.04544}, 2020.

\bibitem[Faroughi et~al.(2023)Faroughi, Darvishi, and Rezaei]{Faroughi2023}
Shirko Faroughi, Ali Darvishi, and Shahed Rezaei.
\newblock On the order of derivation in the training of physics-informed neural
  networks: case studies for non-uniform beam structures.
\newblock \emph{Acta Mechanica}, 234, 2023.

\bibitem[Wang et~al.(2023)Wang, Sankaran, Wang, and
  Perdikaris]{wang2023experts}
Sifan Wang, Shyam Sankaran, Hanwen Wang, and Paris Perdikaris.
\newblock An expert's guide to training physics-informed neural networks.
\newblock \emph{arXiv:2308.08468}, 2023.

\bibitem[Li et~al.(2023)Li, Zheng, Kovachki, Jin, Chen, Liu, Azizzadenesheli,
  and Anandkumar]{li2023physicsinformed}
Zongyi Li, Hongkai Zheng, Nikola Kovachki, David Jin, Haoxuan Chen, Burigede
  Liu, Kamyar Azizzadenesheli, and Anima Anandkumar.
\newblock Physics-informed neural operator for learning partial differential
  equations.
\newblock \emph{arXiv:2111.03794}, 2023.

\bibitem[Wang et~al.(2021)Wang, Wang, and Perdikaris]{Wang_Paris2021}
Sifan Wang, Hanwen Wang, and Paris Perdikaris.
\newblock Learning the solution operator of parametric partial differential
  equations with physics-informed deeponets.
\newblock \emph{Science Advances}, 7\penalty0 (40):\penalty0 eabi8605, 2021.

\bibitem[Zhu et~al.(2019)Zhu, Zabaras, Koutsourelakis, and
  Perdikaris]{ZHU201956}
Yinhao Zhu, Nicholas Zabaras, Phaedon-Stelios Koutsourelakis, and Paris
  Perdikaris.
\newblock Physics-constrained deep learning for high-dimensional surrogate
  modeling and uncertainty quantification without labeled data.
\newblock \emph{Journal of Computational Physics}, 394:\penalty0 56--81, 2019.

\bibitem[Gao et~al.(2021)Gao, Sun, and Wang]{GAO2021110079}
Han Gao, Luning Sun, and Jian-Xun Wang.
\newblock Phygeonet: Physics-informed geometry-adaptive convolutional neural
  networks for solving parameterized steady-state pdes on irregular domain.
\newblock \emph{Journal of Computational Physics}, 428:\penalty0 110079, 2021.

\bibitem[Liu et~al.(2024)Liu, Zhu, Lu, Sun, and Wang]{Liu2024}
Xin-Yang Liu, Min Zhu, Lu~Lu, Hao Sun, and Jian-Xun Wang.
\newblock Multi-resolution partial differential equations preserved learning
  framework for spatiotemporal dynamics.
\newblock \emph{Communications Physics}, 2024.

\bibitem[Zhang and Gu(2021)]{ZHANG2021100220}
Zhizhou Zhang and Grace~X Gu.
\newblock Physics-informed deep learning for digital materials.
\newblock \emph{Theoretical and Applied Mechanics Letters}, 11\penalty0
  (1):\penalty0 100220, 2021.

\bibitem[Kontolati et~al.(2023)Kontolati, Goswami, Karniadakis, and
  Shields]{kontolati2023learning}
Katiana Kontolati, Somdatta Goswami, George~Em Karniadakis, and Michael~D.
  Shields.
\newblock Learning in latent spaces improves the predictive accuracy of deep
  neural operators, 2023.

\bibitem[Zhang and Garikipati(2023)]{ZHANG2023116214}
Xiaoxuan Zhang and Krishna Garikipati.
\newblock Label-free learning of elliptic partial differential equation solvers
  with generalizability across boundary value problems.
\newblock \emph{Computer Methods in Applied Mechanics and Engineering},
  417:\penalty0 116214, 2023.

\bibitem[Ren et~al.(2022)Ren, Rao, Liu, Wang, and Sun]{REN2022114399}
Pu~Ren, Chengping Rao, Yang Liu, Jian-Xun Wang, and Hao Sun.
\newblock Phycrnet: Physics-informed convolutional-recurrent network for
  solving spatiotemporal pdes.
\newblock \emph{Computer Methods in Applied Mechanics and Engineering},
  389:\penalty0 114399, 2022.

\bibitem[Zhao et~al.(2023)Zhao, Gong, Zhang, Yao, and Chen]{ZHAO2023105516}
Xiaoyu Zhao, Zhiqiang Gong, Yunyang Zhang, Wen Yao, and Xiaoqian Chen.
\newblock Physics-informed convolutional neural networks for temperature field
  prediction of heat source layout without labeled data.
\newblock \emph{Engineering Applications of Artificial Intelligence},
  117:\penalty0 105516, 2023.

\bibitem[Zhang et~al.(2023)Zhang, Yan, Liu, Zhang, Han, and
  Wang]{ZHANG2023111919}
Zhao Zhang, Xia Yan, Piyang Liu, Kai Zhang, Renmin Han, and Sheng Wang.
\newblock A physics-informed convolutional neural network for the simulation
  and prediction of two-phase darcy flows in heterogeneous porous media.
\newblock \emph{Journal of Computational Physics}, 477:\penalty0 111919, 2023.

\bibitem[Fuhg et~al.(2023)Fuhg, Karmarkar, Kadeethum, Yoon, and
  Bouklas]{Fuhg2023}
Jan~Niklas Fuhg, Arnav Karmarkar, Teeratorn Kadeethum, Hongkyu Yoon, and
  Nikolaos Bouklas.
\newblock Deep convolutional ritz method: parametric pde surrogates without
  labeled data.
\newblock \emph{Applied Mathematics and Mechanics}, 44\penalty0 (7):\penalty0
  1151--1174, 2023.

\bibitem[He et~al.(2023)He, Abueidda, Koric, and Jasiuk]{He2023}
Junyan He, Diab Abueidda, Seid Koric, and Iwona Jasiuk.
\newblock On the use of graph neural networks and shape-function-based gradient
  computation in the deep energy method.
\newblock \emph{International Journal for Numerical Methods in Engineering},
  124\penalty0 (4):\penalty0 864--879, 2023.

\bibitem[Rezaei et~al.(2024{\natexlab{b}})Rezaei, Moeineddin, Kaliske, and
  Apel]{rezaei2024integration}
Shahed Rezaei, Ahmad Moeineddin, Michael Kaliske, and Markus Apel.
\newblock Integration of physics-informed operator learning and finite element
  method for parametric learning of partial differential equations.
\newblock \emph{arXiv:2401.02363}, 2024{\natexlab{b}}.

\bibitem[Yamazaki et~al.(2024)Yamazaki, Harandi, Muramatsu, Viardin, Apel,
  Brepols, Reese, and Rezaei]{yamazaki2024finite}
Yusuke Yamazaki, Ali Harandi, Mayu Muramatsu, Alexandre Viardin, Markus Apel,
  Tim Brepols, Stefanie Reese, and Shahed Rezaei.
\newblock A finite element-based physics-informed operator learning framework
  for spatiotemporal partial differential equations on arbitrary domains.
\newblock 2024.

\bibitem[Chen and Chen(1995)]{Chen1995UniversalAT}
Tianping Chen and Hong Chen.
\newblock Universal approximation to nonlinear operators by neural networks
  with arbitrary activation functions and its application to dynamical systems.
\newblock \emph{IEEE transactions on neural networks}, 6 4:\penalty0 911--7,
  1995.

\bibitem[Wang et~al.(2022)Wang, Wang, and Perdikaris]{wang2022improved}
Sifan Wang, Hanwen Wang, and Paris Perdikaris.
\newblock Improved architectures and training algorithms for deep operator
  networks.
\newblock \emph{Journal of Scientific Computing}, 92\penalty0 (2):\penalty0 35,
  2022.

\bibitem[Haghighat et~al.(2024)Haghighat, bin Waheed, and
  Karniadakis]{haghighat2024deeponet}
Ehsan Haghighat, Umair bin Waheed, and George Karniadakis.
\newblock En-deeponet: An enrichment approach for enhancing the expressivity of
  neural operators with applications to seismology.
\newblock \emph{Computer Methods in Applied Mechanics and Engineering},
  420:\penalty0 116681, 2024.

\bibitem[Laschet et~al.(2022)Laschet, Abouridouane, Fernández, Budnitzki, and
  Bergs]{LASCHET2022143125}
Gottfried Laschet, M.~Abouridouane, M.~Fernández, M.~Budnitzki, and T.~Bergs.
\newblock Microstructure impact on the machining of two gear steels. part 1:
  Derivation of effective flow curves.
\newblock \emph{Materials Science and Engineering: A}, 845:\penalty0 143125,
  2022.

\bibitem[Vogiatzief et~al.(2022)Vogiatzief, Evirgen, Pedersen, and
  Hecht]{VOGIATZIEF2022165658}
D.~Vogiatzief, A.~Evirgen, M.~Pedersen, and U.~Hecht.
\newblock Laser powder bed fusion of an al-cr-fe-ni high-entropy alloy produced
  by blending of prealloyed and elemental powder: Process parameters,
  microstructures and mechanical properties.
\newblock \emph{Journal of Alloys and Compounds}, 918:\penalty0 165658, 2022.

\bibitem[Mianroodi et~al.(2022)Mianroodi, Rezaei, Siboni, Xu, and
  Raabe]{mianroodi2022lossless}
Jaber~Rezaei Mianroodi, Shahed Rezaei, Nima~H Siboni, Bai-Xiang Xu, and Dierk
  Raabe.
\newblock Lossless multi-scale constitutive elastic relations with artificial
  intelligence.
\newblock \emph{npj Computational Materials}, 8:\penalty0 1--12, 2022.

\bibitem[Bradbury et~al.(2018)Bradbury, Frostig, Hawkins, Johnson, Leary,
  Maclaurin, Necula, Paszke, Vander{P}las, Wanderman-{M}ilne, and
  Zhang]{jax2018github}
James Bradbury, Roy Frostig, Peter Hawkins, Matthew~James Johnson, Chris Leary,
  Dougal Maclaurin, George Necula, Adam Paszke, Jake Vander{P}las, Skye
  Wanderman-{M}ilne, and Qiao Zhang.
\newblock {JAX}: composable transformations of {P}ython+{N}um{P}y programs,
  2018.
\newblock URL \url{http://github.com/google/jax}.

\bibitem[Haghighat and Juanes(2021)]{SciANN}
Ehsan Haghighat and Ruben Juanes.
\newblock Sciann: A keras/tensorflow wrapper for scientific computations and
  physics-informed deep learning using artificial neural networks.
\newblock \emph{Computer Methods in Applied Mechanics and Engineering},
  373:\penalty0 113552, 2021.

\bibitem[Ramachandran et~al.(2017)Ramachandran, Zoph, and
  Le]{ramachandran2017searching}
Prajit Ramachandran, Barret Zoph, and Quoc~V. Le.
\newblock Searching for activation functions.
\newblock \emph{arXiv:1710.05941}, 2017.

\bibitem[Kingma and Ba(2017)]{kingma2017adam}
Diederik~P. Kingma and Jimmy Ba.
\newblock Adam: A method for stochastic optimization.
\newblock \emph{arXiv:1412.6980}, 2017.

\bibitem[Koopas et~al.(2024)Koopas, Rezaei, Rauter, Ostwald, and
  Lammering]{koopas2024introducing}
Rasoul~Najafi Koopas, Shahed Rezaei, Natalie Rauter, Richard Ostwald, and Rolf
  Lammering.
\newblock Introducing a microstructure-embedded autoencoder approach for
  reconstructing high-resolution solution field from reduced parametric space.
\newblock \emph{arXiv preprint arXiv:2405.01975}, 2024.

\bibitem[Tancik et~al.(2020)Tancik, Srinivasan, Mildenhall, Fridovich-Keil,
  Raghavan, Singhal, Ramamoorthi, Barron, and Ng]{tancik2020fourier}
Matthew Tancik, Pratul~P. Srinivasan, Ben Mildenhall, Sara Fridovich-Keil,
  Nithin Raghavan, Utkarsh Singhal, Ravi Ramamoorthi, Jonathan~T. Barron, and
  Ren Ng.
\newblock Fourier features let networks learn high frequency functions in low
  dimensional domains.
\newblock 2020.

\bibitem[Harandi et~al.(2024)Harandi, Moeineddin, Kaliske, Reese, and
  Rezaei]{Harandi2023}
Ali Harandi, Ahmad Moeineddin, Michael Kaliske, Stefanie Reese, and Shahed
  Rezaei.
\newblock Mixed formulation of physics-informed neural networks for
  thermo-mechanically coupled systems and heterogeneous domains.
\newblock \emph{International Journal for Numerical Methods in Engineering},
  125\penalty0 (4):\penalty0 e7388, 2024.

\bibitem[Taylor(2014)]{taylor_feap_2014}
R.~L. Taylor.
\newblock {FEAP} - finite element analysis program, 2014.

\bibitem[Geuzaine and Remacle(2009)]{Gmsh}
Christophe Geuzaine and Jean-François Remacle.
\newblock Gmsh: A 3-d finite element mesh generator with built-in pre- and
  post-processing facilities, 2009.

\bibitem[Bathe(1996)]{bathe}
Klaus-Jurgen Bathe.
\newblock \emph{Finite Element Procedures}.
\newblock Prentice Hall, Englewood Cliffs, NJ, 1st edition, 1996.

\bibitem[Hughes(1987)]{hughes}
Thomas J.~R. Hughes.
\newblock \emph{The Finite Element Method: Linear Static and Dynamic Finite
  Element Analysis}.
\newblock Prentice Hall, Englewood Cliffs, NJ, 1st edition, 1987.

\bibitem[Boull{\'e} and Townsend(2023)]{boulle2023}
Nicolas Boull{\'e} and Alex Townsend.
\newblock A mathematical guide to operator learning.
\newblock \emph{arXiv:2312.14688}, 2023.

\end{thebibliography}

\end{document}